\documentclass{article}

\usepackage{iclr2025_conference,times}

\newcommand{\system}{\textsc{AndroidWorld}\xspace}
\title{\system: A Dynamic Benchmarking \\Environment for Autonomous Agents}

\usepackage{hyperref}
\usepackage{url}
\usepackage{listings}
\usepackage{graphicx}
\usepackage{xspace}
\usepackage{xcolor}
\usepackage{array}
\usepackage{booktabs}
\usepackage{longtable}
 \usepackage{ragged2e}
 \usepackage{seqsplit}

\usepackage{tabularx}
\usepackage{amssymb}
\usepackage{amsmath}
\usepackage{pifont}
\usepackage{caption} 
\usepackage{subcaption}
\usepackage{wrapfig}

\usepackage[affil-it]{authblk}
\author[1]{Christopher Rawles$^*$}
\author[2]{Sarah Clinckemaillie$^{\dagger}$}
\author[2]{Yifan Chang$^{\dagger}$}
\author[2]{Jonathan Waltz}
\author[2]{Gabrielle Lau}
\author[2]{Marybeth Fair}
\author[1]{Alice Li}
\author[1]{William Bishop}
\author[1]{Wei Li}
\author[1]{Folawiyo Campbell-Ajala}
\author[1]{Daniel Toyama}
\author[1]{Robert Berry}
\author[2]{Divya Tyamagundlu}
\author[1]{Timothy Lillicrap}
\author[1]{Oriana Riva}

\affil[1]{Google DeepMind}
\affil[2]{Google}

\definecolor{codegreen}{rgb}{0,0.6,0}
\definecolor{codegray}{rgb}{0.5,0.5,0.5}
\definecolor{codepurple}{rgb}{0.58,0,0.82}
\definecolor{backcolour}{rgb}{0.95,0.95,0.92}

\lstdefinestyle{mystyle}{
    backgroundcolor=\color{backcolour},   
    commentstyle=\color{codegreen},
    keywordstyle=\color{magenta},
    numberstyle=\tiny\color{codegray},
    stringstyle=\color{codepurple},
    basicstyle=\tiny\tt,
    breakatwhitespace=false,         
    breaklines=true,                 
    captionpos=b,                    
    keepspaces=true,                 
    numbers=left,                    
    numbersep=5pt,                  
    showspaces=false,                
    showstringspaces=false,
    showtabs=false,                  
    tabsize=2
}

\newcommand{\agent}{\textsc{M3A}\xspace}
\newcommand{\agentsimple}{\textsc{M3A-Simple}\xspace}
\newcommand{\ntasks}{116\xspace}
\newcommand{\napps}{20\xspace}
\newcommand{\miniwob}{MiniWoB++\xspace}
\newcommand{\mobileminiwob}{MobileMiniWoB++\xspace}

\newcommand{\humanresult}{80.0}
\newcommand{\mthreearesult}{30.6}

\newcommand{\cmark}{\textcolor{green}{\ding{51}}} 
\newcommand{\xmark}{\textcolor{red}{\ding{55}}} 

\newcommand{\location}{\url{https://github.com/google-research/android_world}}

\usepackage[title]{appendix}
\hypersetup{
 colorlinks=true,
 linkcolor=blue,
 filecolor=magenta,      
 urlcolor=blue,
}

\begin{document}

\iclrfinalcopy 

\begingroup
\renewcommand{\thefootnote}{\fnsymbol{footnote}}
\footnotetext[1]{Lead contributor. Contact: \texttt{crawles@google.com}}
\footnotetext[2]{Equal contribution.}
\endgroup

\maketitle

\begin{abstract}

Autonomous agents that execute human tasks by controlling computers can enhance human productivity and application accessibility. However, progress in this field will be driven by realistic and reproducible benchmarks. We present \system, a fully functional Android environment that provides reward signals for \ntasks programmatic tasks across \napps real-world Android apps. Unlike existing interactive environments, which provide a static test set, \system dynamically constructs tasks that are parameterized and expressed in natural language in unlimited ways, thus enabling testing on a much larger and more realistic suite of tasks. To ensure reproducibility, each task includes dedicated initialization, success-checking, and tear-down logic, which modifies and inspects the device's system state.

We experiment with baseline agents to test \system and provide initial results on the benchmark. Our best agent can complete \mthreearesult\% of \system's tasks, leaving ample room for future work. Furthermore, we adapt a popular desktop web agent to work on Android, which we find to be less effective on mobile, suggesting future research is needed to achieve universal, cross-platform agents. Finally, we also conduct a robustness analysis, showing that task variations can significantly affect agent performance, demonstrating that without such testing, agent performance metrics may not fully reflect practical challenges. \system and the experiments in this paper are available at \location.
\end{abstract}

\section{Introduction}
\label{sec:intro}

Autonomous agents that interpret natural language instructions and operate computing devices can provide enormous value to users by automating repetitive tasks, augmenting human intelligence, and accomplishing complex workflows.  However, a key research challenge remains the realistic evaluation of these agents in real-world settings. Despite growing enthusiasm for building autonomous agents~\citep{deng2023mind2web,rawles2023android,zheng2023seeact,koh2024visualwebarena,kim2024language,he2024webvoyager, AutoGPT, wu2023autogen, OpenAgents} most existing approaches for evaluation compare an agent's actions at each step to a previously collected human demonstration~\citep{deng2023mind2web, rawles2023android, yang2023appagent, zhang2023look, Lu2024-pg, zhang2024android, yan2023gpt4v,li2024-android-control}.  Measuring performance in this way can be misleading because when performing tasks online in real environments agents can take multiple paths to solve tasks, environments may behave non-deterministically, and agents can dynamically learn from mistakes to correct their actions~\citep{shinn2023reflexion, liu2018, Li2023-vo, Pan2024-kc}. For this reason, online evaluation of agents in realistic environments able to reward task outcome provides a gold standard for evaluation.  While there is an emerging body of work to address this need across different environments~\citep{zhou2023webarena, koh2024visualwebarena, drouin2024workarena, lee2024benchmarking,  xie2024osworld,bonatti2024windowsagentarenaevaluating, zheng2024agentstudiotoolkitbuildinggeneral}, there is no comprehensive solution for mobile platforms, such as Android, which are used by billions of users and therefore represent environments in which automation agents may be very productively employed. We introduce \system to address this.

At its core, \system offers a reliable means of obtaining reward signals for tasks performed by agents in realistic mobile environments. Reward signals are quantitative metrics that indicate functional correctness of a task, i.e. is the stated goal achieved? For example, for the task ``Send a text message to Jane confirming I'll be there," a positive reward indicates that the relevant message has been sent. Unlike simulated environments \citep{Tassa2018-kg, shridhar2020alfred} or games \citep{mnih2013playing, silver16, Vinyals2019-tq, wang2023voyager, weihao2024cradle, android_env}, real-world apps and websites do not inherently offer explicit reward signals. While human \citep{rawles2023android, zheng2023seeact, Pan2024-kc, kinniment2023evaluating} or LLM-based \citep{chiang2024chatbot, Zheng2023-ds, liu2023geval, Du2023-rw, ma2023eureka, Pan2024-kc, he2024webvoyager} judges can be employed to reward the outcome of a task, these approaches scale poorly or are not fully reliable, respectively. Alternatively, environments for autonomous agents which provide automated ground-truth rewards for complex workflows have been developed ~\citep{yao2023webshop,zhou2023webarena,koh2024visualwebarena,xie2024osworld,bonatti2024windowsagentarenaevaluating}. We find two problems with these environments. First, they are constrained to desktop computing environments, overlooking the mobile domain, which is of paramount importance given the ubiquity and diversity of mobile devices in the real world. Secondly, they are limited in their real-world diversity and scale. Crucially, unlike in real-world scenarios where conditions and task inputs vary widely, these environments support only static test specifications, meaning that when task parameters deviate, the reward signal is likely to break.

\begin{figure}[t]
\centering
\centering
\includegraphics[width=0.99\textwidth]{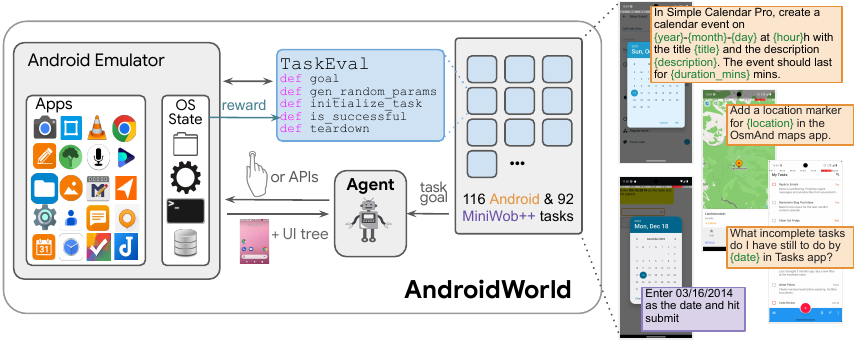}
\label{fig:AndroidWorld}
\caption{\system is an environment for building and testing autonomous agents.}
\label{fig:architecture}
\end{figure}

We seek to develop a comprehensive benchmark that addresses the limitations of the existing approaches above for evaluating automation agents in mobile environments. \system does this by spanning \napps Android apps on a total of \ntasks programmatic tasks to provide ground truth-rewards. Unlike existing test environments (MiniWoB++~\citep{miniwob} being a notable exception), each task in \system is dynamically instantiated using randomly-generated parameters, challenging agents with millions of unique task goals and conditions. While MiniWob++ consists of simple, synthetic websites, \system leverages actual Android applications. A main challenge that \system must address is how to ensure that reward signals are durable when using real-world applications and varying task parameters dynamically. \system's solves this by leveraging the extensive and consistent state management capabilities of the Android OS, using the same mechanisms that the apps themselves utilize to store and update data.

In addition to providing a comprehensive benchmark, \system is lightweight, requiring only 2 GB of memory and 8 GB of disk space, and is designed with convenience in mind. It connects agents to the Android OS by leveraging the Python library AndroidEnv \citep{android_env} to connect to the freely available Android Emulator.\footnote{The Android Emulator is packaged as part of Android Studio, which can be downloaded from https://developer.android.com/studio} In addition to the \ntasks Android tasks, we extend \system with web tasks by integrating the \miniwob \citep{miniwob, web-workflows18} benchmark into it.

To demonstrate \system's usefulness as a benchmark, we build and release a multi-modal agent, \emph{\agent} (Multimodal Autonomous Agent for Android), and establish state-of-the-art results on \system. We analyze \agent's performance using both multimodal and text-only input, and we observe that while multimodal perception can improve performance in some cases, it generally does not outperform the text-only approach. On \system, \agent achieves a \mthreearesult\% success rate, which surpasses that of a web agent adapted for Android but remains significantly lower than the human success rate of \humanresult\%. In pursuit of building robust UI control agents, our study includes comprehensive tests under varied real-world conditions, demonstrating significant performance variations primarily driven by changes in intent parameters.

We make the following contributions: (i) the creation of a new, highly diverse and realistic mobile UI control agent environment; (ii) establishment of benchmark performance with a state-of-the-art multimodal agent, and (iii) a careful analysis demonstrating the need to evaluate agents across variable task parameters and conditions due to the inherent stochasticity in both models and environments.

\section{Related Work}
\label{sec:related_work}

\begin{table}[t]
\centering
\caption{Comparison of different datasets and environments for benchmarking computer agents.}
\label{tab:summary_comparsion}
\resizebox{\textwidth}{!}{%
\begin{tabular}{@{}lccccll@{}}
\toprule
& Env? & \# of apps & \# task  & Avg \# task &  \multicolumn{1}{c}{Reward} & Platform \\ 
& & or websites & templates & instances & \multicolumn{1}{c}{method} & \\
\midrule
\textsc{GAIA} & \xmark & n/a & ~~466 & 1 & text-match & None \\
\textsc{Mind2Web} & \xmark & 137 & 2350 & 1 & None & Desktop Web \\
\textsc{WebLINX} & \xmark & 155 & 2337 & 1 & None & Desktop Web \\
\textsc{WebVoyager} & \xmark & ~15 & ~~643 & 1 & LLM judge & Desktop Web \\
\textsc{PixelHelp} & \xmark & ~~~4 & ~~187 & 1 & None & Android \\
\textsc{MetaGUI} & \xmark & ~~~6 & 1125 & 1 & None & Android \\
\textsc{MoTiF} & \xmark & 125 & 4707 & 1 & None & Android (Apps+Web) \\
\textsc{AitW} & \xmark & 357+~ & 30378~~ & 1 & None & Android (Apps+Web) \\
\textsc{AndroidControl} & \xmark & 833~ & 15283~~ & 1 & None & Android (Apps+Web) \\
\textsc{OmniAct} & \xmark & ~60+~ & 9802 & 1 & None & Desktop (Apps+Web) \\
\textsc{AndroidArena} & \xmark & ~13 & ~~221 & 1 & Action match/LLM & Android (Apps+Web) \\
\textsc{LLamaTouch} & \xmark & ~57 & ~~496 & 1 & Screen match & Android (Apps+Web) \\
\midrule
\textsc{MiniWoB++} & \cmark & ~~~1 & ~114 & $\infty$ & HTML/JS state & Web (synthetic) \\
\textsc{WebShop} & \cmark & ~~~1 & 12k & 1 & product attrs match & Desktop Web \\
\textsc{WebArena} & \cmark & ~~~6 & ~241 & ~~~3.3 & url/text-match & Desktop Web \\
\textsc{VisualWebArena} & \cmark & ~~~4    & ~314 & ~~~2.9 & url/text/image-match & Desktop Web \\
\textsc{WorkArena} & \cmark & ~~~1    & ~~~29 & 622.4~ & cloud state & Desktop Web \\
\textsc{Mobile-Env} & \cmark & ~~~1    & ~~~13 & ~11.5 & regex & Android (Apps) \\
\textsc{B-MoCA} & \cmark & ~~~4    & ~~~~~6 & ~~~1.9 & regex & Android (Apps+Web) \\
\textsc{MMInA} & \cmark & ~14 & 1050 & 1 & text-match & Desktop web \\
\textsc{OSWorld} & \cmark & ~~~9 & ~369 & 1 & device/cloud state & Desktop (Apps+Web) \\
\textsc{WindowsAgentArena} & \cmark & ~~11 & ~154 & 1 & device state & Desktop (Apps+Web) \\
\textsc{AgentStudio} & \cmark & ~~9 & ~205 & 1 & device state & Desktop (Apps+Web) \\
\midrule
\textbf{\system} & \cmark & ~\napps & ~\ntasks & $\infty$ & device state & Android (Apps+Web) \\
\bottomrule
\end{tabular}%
}
\end{table}

Table~\ref{tab:summary_comparsion} compares existing evaluation environments for autonomous UI agents. 

\subsection{Interactive evaluation environments}

Effective evaluation of autonomous agents requires benchmarks that mimic real-world scenarios, but also interactive environments that provide reward signals upon successful task completion \citep{rawles2023android, deng2023mind2web, abramson2022evaluating, Ruan2023-gu-toolemu, Chen2021-xc}. Many existing benchmarking environments target web browsing. MiniWoB++ \citep{miniwob, liu2018} consists of small, synthetic HTML pages with parameterizable tasks which allow for unlimited task variability.
WebShop \citep{yao2023webshop} provides a simulated e-commerce environment, whereas WebArena \citep{zhou2023webarena} and VisualWebArena \citep{koh2024visualwebarena} consist of simulated websites across up to six domains. WorkArena \citep{drouin2024workarena} consists of 29 tasks for enterprise software.  GAIA \citep{Mialon2023-ut} is a static dataset that tests an agent's ability to interact with live web environments. MMInA \citep{zhang2024mmina} is a multihop and multimodal benchmark designed to evaluate agents for compositional Internet tasks.

Towards building computer use agents, OSWorld \citep{xie2024osworld}, WindowsAgentArena~\citep{bonatti2024windowsagentarenaevaluating}, and AgentStudio~\citep{zheng2024agentstudiotoolkitbuildinggeneral} provide a test suite of tasks for desktop computer interfaces and custom execution-based evaluation scripts across 9, 11, and 9 apps, respectively. In the mobile domain, existing benchmarks are limited and do not capture the diversity of real-world mobile interactions, containing low-complexity tasks or on a limited number of applications. B-MoCA's \citep{lee2024benchmarking} evaluation is based on 6 simple tasks (e.g., "Call 911", "turn on airplane mode") across 4 apps\footnote{Based on what reported in the Experiments Section of the B-MoCA manuscript as of October 1\textsuperscript{st}, 2024.}, validated using regular expressions. Mobile-Env~\citep{mobile-env} offers task reproducibility limited to 13 task templates for a single app (WikiHow).

While \system shares the mobile OS focus of B-MoCA and Mobile-Env, it is more comparable to OSWorld (and WindowsAgentArena, which builds on top of OSWorld) in terms of task complexity and the diversity of interactions it supports. \system enhances OSWorld's approach by dynamically constructing the start states of an agent's run and varying the task parameters in unlimited ways, thus allowing for a new type of evaluation under varying real-world conditions.

Other studies leverage human evaluation \citep{rawles2023android, zheng2023seeact,bishop2024latent} for tasks where automatic evaluation is not available.  Lastly, emerging research \citep{Pan2024-kc, he2024webvoyager, android-arena, zheng2024agentstudiotoolkitbuildinggeneral} explores the potential of multimodal models to generalize agent evaluations to new settings, though this area requires further research to achieve accuracy comparable to manually-coded rewards.

AndroidEnv~\citep{android_env} provides a mechanism to manage communication with the Android emulator, similar to Playwright and Selenium for web environments. While \system leverages this functionality, it diverges in its reward system. AndroidEnv's approach requires modifying application source code and implementing task-specific logging statements, making it well-suited for gaming environments with easily verifiable success criteria. In contrast, \system implements a non-invasive reward mechanism, allowing it to create a benchmark suite for apps whose source code is unavailable and to reuse validation components across different apps. This approach enables \system to cover a broader range of real-world mobile tasks.

\subsection{Static datasets for UI automation}

Datasets derived from human interactions provide proxy metrics that correlate with real-world agent performance \citep{li-acl20, motif, deng2023mind2web, rawles2023android}. On mobile platforms, AitW \citep{rawles2023android}, AndroidControl~\citep{li2024-android-control}, PixelHelp \citep{li-acl20}, AndroidArena \citep{android-arena}, LlamaTouch \citep{llama-touch}, UGIF \citep{ugif}, and MoTIF \citep{motif} consist of demonstrations across Android apps and mobile websites, with screens often represented via accessibility trees. In contrast, desktop web environments typically utilize the DOM for representing website content, with Mind2Web \citep{deng2023mind2web}, OmniAct \citep{kapoor2024omniact} and others, across various desktop websites. Mobile-based datasets frequently involve more complex actions, such as scrolling, which are not as useful in DOM-based desktop interactions where the entire action space is readily accessible. Additionally, API-centric datasets like API-Bank \citep{Li2023-tn}, ToolTalk \citep{farn2023tooltalk}, and ToolBench \citep{Xu2023-uw} assess agents' capabilities to manipulate computer systems via APIs.

\subsection{Interactive agents}

Prior to today's foundation models, traditional approaches to developing user interface-operating agents primarily used reinforcement learning and behavioral cloning to simulate interactions like mouse clicks and keyboard typing \citep{liu2018, li-acl20, appbuddy, gur2022environment, pmlr-v162-humphreys22a}. More recent work leverages off-the-shelf foundational models \citep{geminiteam2023gemini, openai2023gpt4, touvron2023llama} with in-context learning (ICL) and fine-tuning applied to mobile \citep{rawles2023android, hong2023cogagent, wang:chi2023, yan2023gpt4v, zhang2023look, bishop2024latent, zhang2023igniting}, desktop web \citep{zheng2023seeact, deng2023mind2web, zhou2023webarena, koh2024visualwebarena, cheng2024seeclick, lai2024autowebglm, you2024ferretuigroundedmobileui}, and desktop OS \citep{wu2024copilot, zhang2024ufo, xie2024osworld}. Recent work explores agents that reflect on system state \citep{shinn2023reflexion, Yao2022-nv, madaan2024self} by leveraging exploration, self-evaluation, and retry-capabilities for continual learning and adaptation \citep{Li2023-vo,yang2023appagent,Pan2024-kc,wu2024copilot,gao2023assistgui,murty2024bagel}.

\section{\system}
\label{sec:androidworld}

\subsection{Android for autonomous agents}

Android is an ideal environment for developing autonomous agents. It is the most widely-used OS globally\footnote{\url{https://gs.statcounter.com/os-market-share}} and is highly flexible for research, while providing an open world of the Web\footnote{Mobile is the most popular platform for accessing the web; \url{https://gs.statcounter.com/platform-market-share/desktop-mobile/worldwide/}} and over 2M apps for agents to operate in. Using emulation, an Android environment is easy to deploy, does not require specialized hardware, and can be run on a laptop. Android Virtual Devices or emulator images are well suited for research as they are self-contained, easy to distribute, and configurable.

Compared to desktops, mobile environments like Android present unique challenges for computer-use agents. While mobile UIs are simpler due to smaller screens, their action space is more complex, requiring intricate gestures (e.g., navigating carousels, long-pressing, multi-finger zooming) and often more steps to complete tasks. Unlike web-browser-only environments, Android, as an OS, offers greater flexibility, including function-calling APIs (e.g., sending texts) alongside standard UI actions (click, scroll, type).
\vspace{-.5em}
\subsection{The observation and action space}
\system provides an interface for agents to receive observations and execute actions on Android. It uses AndroidEnv \citep{android_env} and the Android Device Bridge to facilitate interaction between Android and the agent. The observation space consists of a full-resolution screenshot and a UI tree representation developed for accessibility purposes. The action space is similar to that which humans use, consisting of gestures (i.e., tapping, swiping), typing, and navigation buttons (i.e., go home and go back). In addition to these naturalistic actions, \system exposes a limited set of function calling APIs, such as \texttt{send\_text\_message}, to help agents accomplish goals. Appendix~\ref{sec:appendix_environment} provides more details on the observation format and action space.

\subsection{Reproducible and parameterized tasks}

\begin{figure}[t]
  \begin{subfigure}[b]{0.3\textwidth}
    \centering
    \includegraphics[width=\textwidth]{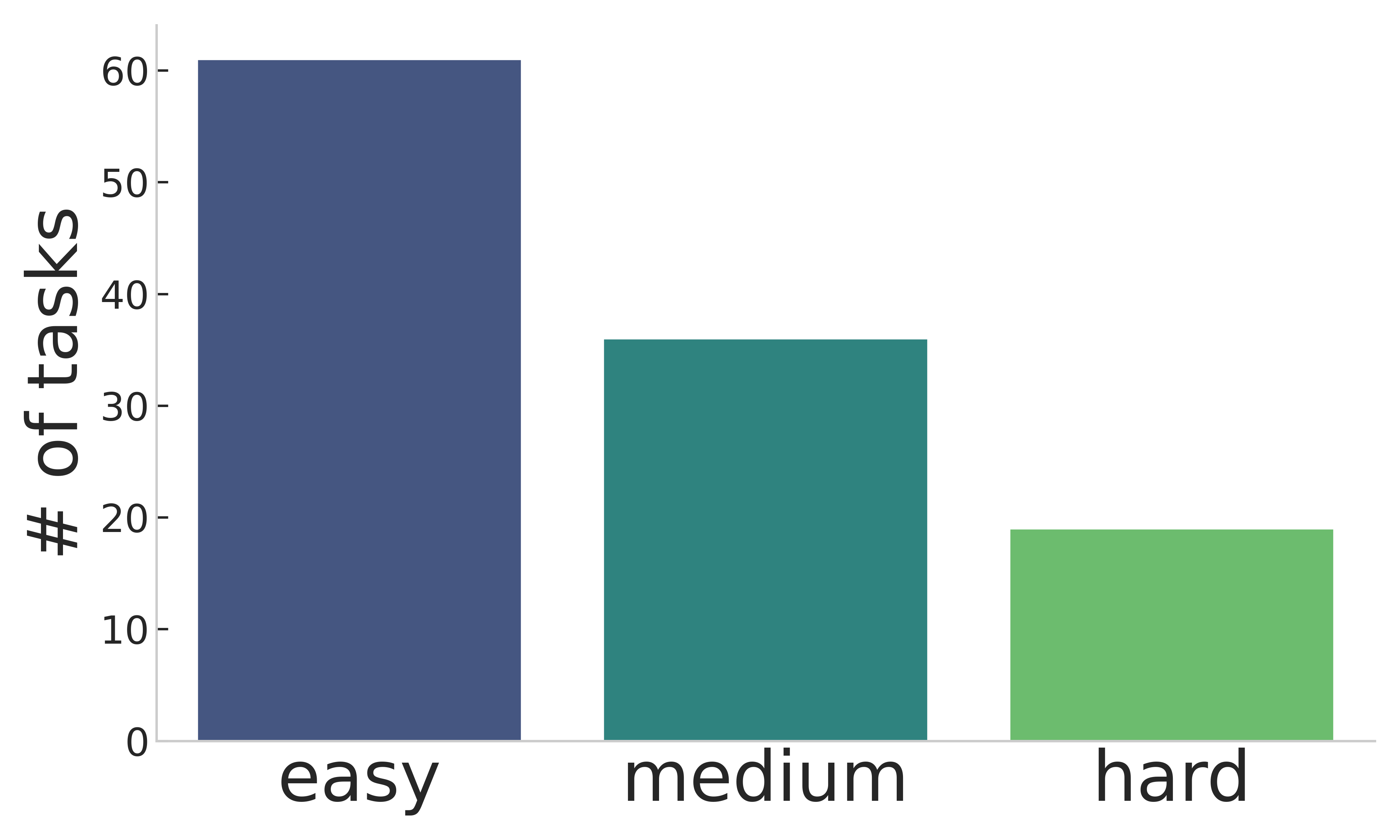}
    \caption{}
    \label{fig:tag1}
  \end{subfigure}
  \hfill
    \begin{subfigure}[b]{0.31\textwidth}
    \centering
    \includegraphics[width=\textwidth]{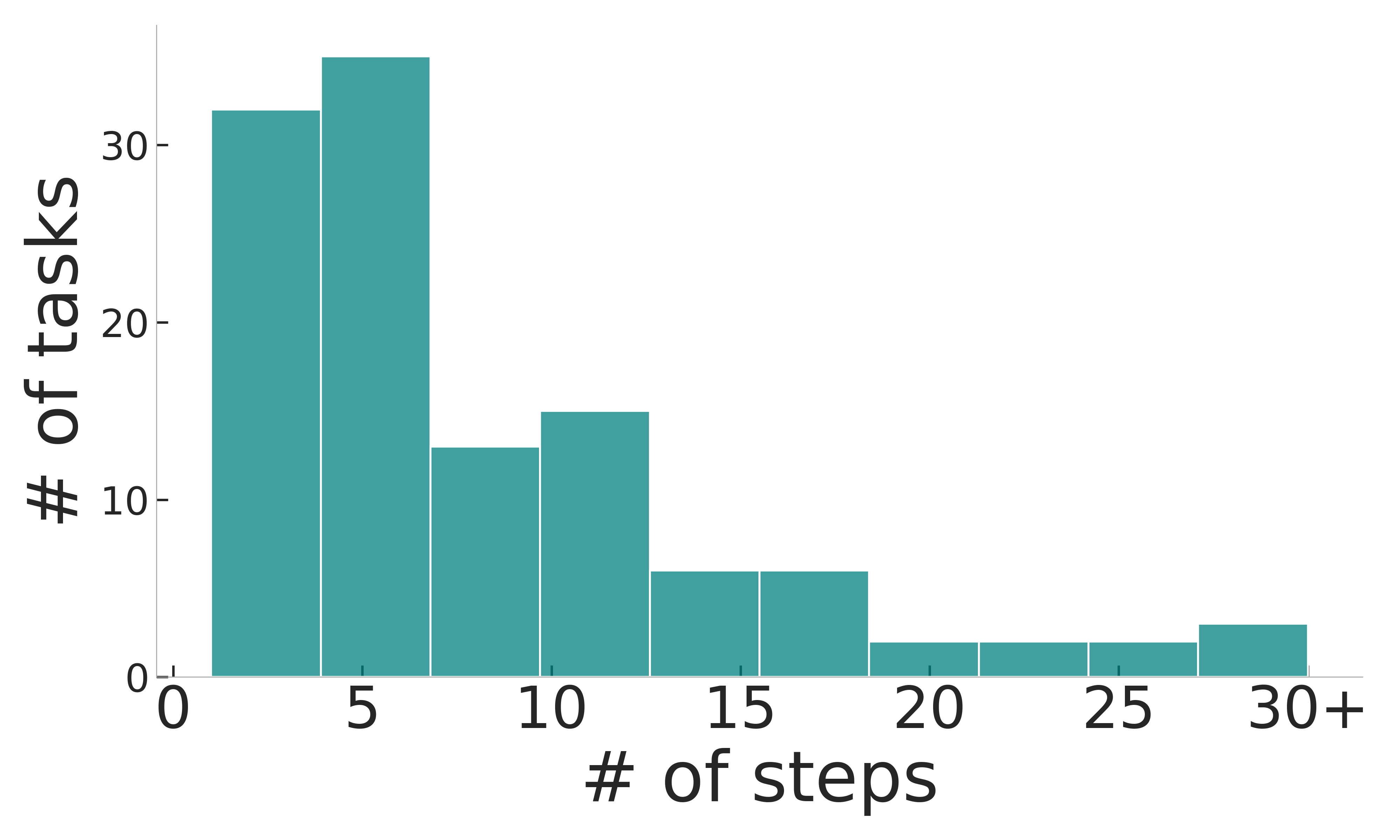}
    \caption{}
    \label{fig:tag2}
  \end{subfigure}
  \hfill
    \begin{subfigure}[b]{0.35\textwidth}
    \centering
    \includegraphics[width=\textwidth]{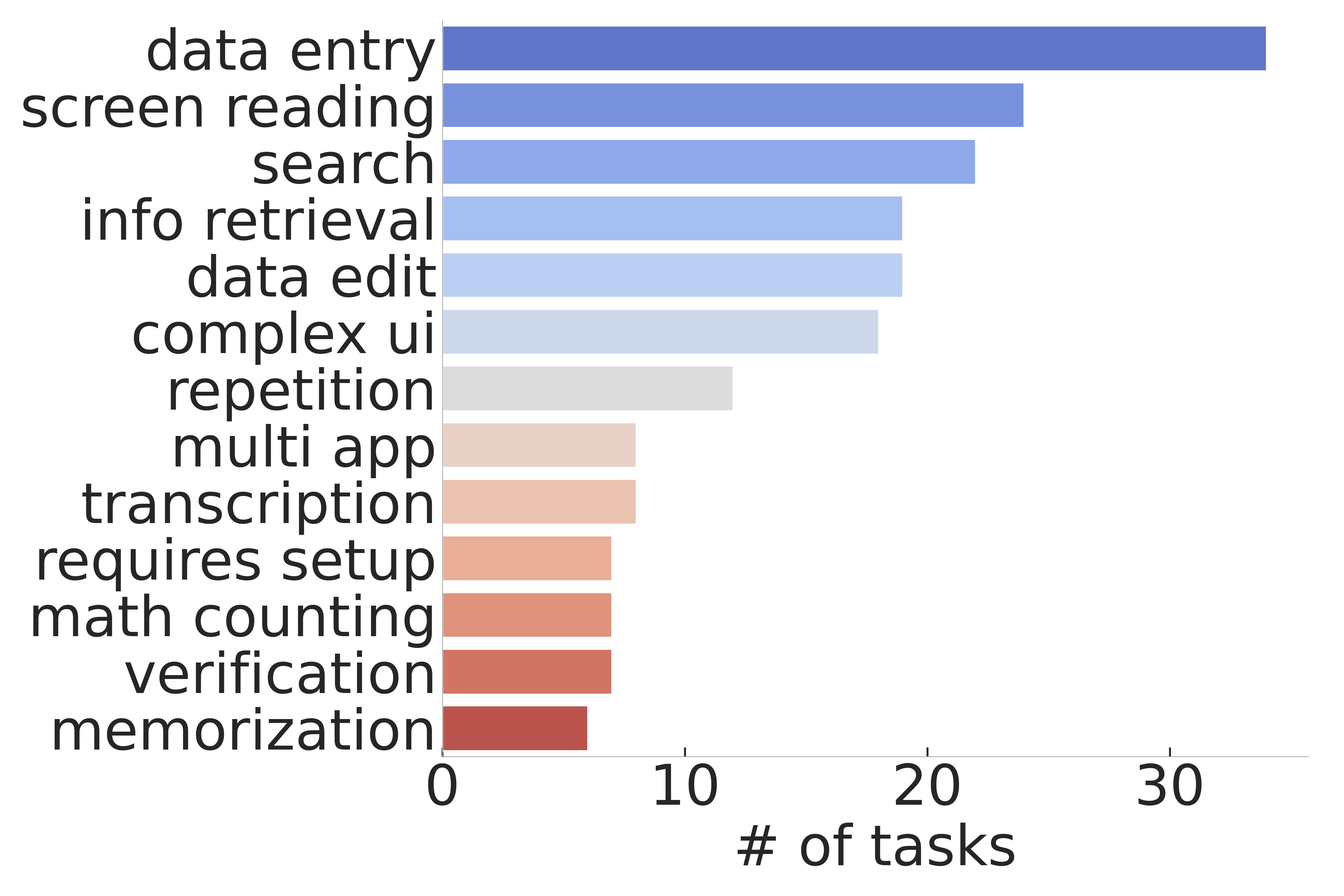}
    \caption{}
    \label{fig:tag3}
  \end{subfigure}
  \caption{Annotators performed the tasks assigned to them, assigned a difficulty level (\ref{fig:tag1}) and estimated the number of steps required to complete each task (\ref{fig:tag2}), using the action space available to an agent. For each task, they selected relevant category tags from a predefined list (\ref{fig:tag3}).}
  \label{fig:tags}
\end{figure}

\system consists of a suite of \ntasks tasks, spread across \napps diverse applications (see Appendix~\ref{sec:benchmark_details} for more details). These tasks simulate practical, everyday activities, including note-taking, scheduling appointments, communicating through messaging, and interacting with system utilities. The suite consists of open-source apps and built-in Android system apps, such as Settings and Contacts. As rated by humans, the tasks vary in difficulty, duration, and categories (Figure \ref{fig:tags}).

To achieve a high degree of reproducibility in real-world scenarios, \system precisely controls the OS and app states in several ways. The Android OS is fixed, consisting of a Pixel 6 emulator running Android 13. At the start of each task, \system resets the device timestamp to October 15th, 2023 at 15:34 UTC, ensuring consistent time-dependent behaviors across all executions. All applications in \system are fully-functional and consists of both open-source apps and OS-level apps included with Android. For the open-source apps, \system maintains a constant environment by installing a fixed version of each app, acquired from F-Droid.\footnote{\url{https://f-droid.org/}}
OS-level apps' versions are determined by the Android OS, which is also fixed. To maintain a reproducible environment, \system utilizes apps that do not require login/authentication and can store their application data on device. 

In addition to managing the states of apps and operating systems, \system precisely defines and controls the state during task execution. Each task has its own unique setup, reward determination logic, and teardown procedures (see Appendix~\ref{sec:task_definition} and \ref{sec:info_retrieval_details} for more details), ensuring a fully reproducible suite of tasks. 

Automatic task \textit{parameterization} is a critical mechanism, unique to \system, to evaluate agents on a much larger and more realistic suite of tasks than current benchmarks support. Achieving this requires significantly more effort than randomly generating new task parameters because it involves developing evaluation logic that remains valid across different task instantiations. It is exactly through its careful state management that in addition to reproducibility AndroidWorld ensures that the reward mechanisms function correctly. Task parameters, initialized randomly at the start of each task based on a controlled random seed, dictate the initial state and influence reward outcomes. Similar to \miniwob \citep{miniwob, web-workflows18}, \system consists of a practically infinite set of varying initial conditions and success criteria. 

This approach enables finer-grained analyses of agent adaptability, essential for real-world deployment. Beyond robustness testing, dynamic task construction supports online learning, particularly reinforcement learning \citep{miniwob, web-workflows18, pmlr-v162-humphreys22a, gur2022environment}, while also streamlining train/test dataset generation for supervised learning \citep{pmlr-v162-humphreys22a, Shaw2023-tj, Furuta2023-ue}.
\begin{table}[t]
\centering
\caption{Selected tasks with code describing validation logic.}
\label{tab:tasks_example_code}
\scalebox{0.72}{
\begin{tabular}{>{\raggedright\arraybackslash}p{0.94\textwidth} >{\raggedright\arraybackslash}p{0.42\textwidth}}
\toprule
\textbf{Task} & \textbf{Validation code} \\
\midrule
In Simple Calendar Pro, create a calendar event on \texttt{\{event.year\}}-\texttt{\{event.month\}}-\texttt{\{event.day\}} at \texttt{\{event.hour\}}h with the title `\texttt{\{event.title\}}' and the description `\texttt{\{event.description\}}'. The event should last for \texttt{\{event.duration\}} mins. & \texttt{event\_exists(event)} \\
\midrule

Send a text message to \texttt{\{phone\_number\}} with message: \texttt{\{message\}}. & \texttt{message\_exists(phone\_number, message, messaging\_db)} \\
\midrule
Create a new drawing in Simple Draw Pro. Name it \texttt{\{file\_name\}}. Save it in the Pictures folder. & \texttt{file\_exists(file\_path)} \\
\midrule
Create a timer with \texttt{\{hours\}} hours, \texttt{\{minutes\}} minutes, and \texttt{\{seconds\}} seconds. Do not start the timer. & \texttt{timer\_displays(time, ui\_hierarchy)} \\
\midrule
Create a new note in Markor named \texttt{\{file\_name\}} with the following text: \texttt{\{text\}}. Share the entire content of the note with the phone number \texttt{\{number\}} via SMS. & \texttt{(file\_exists(file\_name, content=text) + message\_exists(phone\_number, message)) / 2.0} \\
\midrule
Turn on WiFi and open \texttt{\{app\_name\}}. & \texttt{(wifi\_enabled() + app\_launched(app\_name))/2.0} \\
\bottomrule
\end{tabular}
}
\end{table}

\subsection{Durable rewards from system state}

\system provides reward signals primarily by managing application state using the Android Debug Bridge (\texttt{adb}), while also incorporating UI element validation where appropriate. With \texttt{adb}, \system has complete access to system resources including the file system, application databases, and system settings. For tasks where system state inspection is impractical, \system validates task completion by examining UI elements on screen. Determining reward signals from system state has several benefits. It is highly accurate because an application's state can be quickly inspected and manipulated using the same mechanisms that the app itself utilizes. Using the underlying system state is much more durable than matching superficial UI changes. Additionally, it facilitates easy re-use across disparate apps, which tend to use the same underlying caching mechanisms. For instance, logic for checking existence of a specific file is used across many unrelated applications, including those for file management, note-taking, and media playback. For applications leveraging SQLite databases, a common pattern, \system implements evaluators that verify the existence of new and deleted rows. Table~\ref{tab:tasks_example_code} shows examples of the validators in \system. See Table~\ref{tab:all-task-list} for a comprehensive list of all tasks in the suite. Table~\ref{table:task-list} provides selected examples with additional implementation details.

\subsection{Task composability}
Inferring task success from system state enables accurate, reusable evaluations and simplifies creating \textit{composite} tasks by combining existing ones. For instance, ``Create a calendar event with {details} and text the details to {contact}” merges two standalone tasks, facilitated by hermetic initialization and success detection. Composite tasks are more challenging due to their complexity but provide partial rewards for subtask completion, aiding hill climbing. The last two rows of Table \ref{tab:tasks_example_code} show validation code for composite tasks.

\subsection{Integrating MiniWob++}

We implement MiniWoB++ in the \system framework and term it \mobileminiwob. Each \mobileminiwob task is instantiated using the standard \system interface, inheriting from \texttt{TaskEval} base class, and contains methods like \texttt{initialize\_state} and \texttt{is\_successful}. Since MiniWoB++ leverages JavaScript for task configuration and success detection, we built a WebView app to communicate between Python and the app.

\mobileminiwob introduces modifications in both observations and actions compared to the original benchmark. For example, HTML5 \lstinline|<input>| elements are rendered with native Android UI widgets like the date-picker (see Figure~ \ref{fig:miniwob_android_widget}), enhancing the realism of the tasks. \mobileminiwob uses the same observation space as the Android tasks (accessibility tree and screenshot). Notably, it does not include the DOM as in the original implementation. The action space from \system is retained. We manually review and test each task to ensure they are solvable. We excluded twelve of the original tasks that failed to render correctly on Android, presented compatibility issues with the touch interface, or required near real-time interaction, which poses challenges on emulators. Overall, \system supports 92 \miniwob tasks. See Appendix \ref{sec:miniwob_details} for more details.

\section{\system as a computer-control benchmark}
\label{sec:experiments}

To test \system's applicability for autonomous agents, we develop and test a state-of-the-art agent and its variants across all \napps apps and \ntasks tasks, as well as on \mobileminiwob.

\subsection{computer use agents}
\subsubsection{\agent}

We develop a multimodal autonomous agent for Android, \agent. It is zero-shot, integrating ReAct-style \citep{Yao2022-nv} and Reflexion-style \citep{shinn2023reflexion} prompting to consume user instructions and screen content, reason, take actions, and update its decision-making based on the outcome of its actions.

In the first stage, \agent generates an action, represented in JSON, and reasoning for that action. To generate this output, the agent is provided with a list of available action types, guidelines for operating the phone, and a list of UI elements derived from the Android accessibility tree's leaf nodes. The agent receives the current screenshot and a Set-of-Mark (SoM) \citep{yang2023set} annotated screenshot, which includes bounding boxes with numeric labels on the top-left corner for each UI element (see screenshot in Figure~\ref{fig:som_example}). The agent attempts to execute outputted action by referencing the specific mark (if applicable). In addition to the multimodal agent, we have developed a text-only variant that consumes the screen represented using the accessibility tree and selects the relevant action in JSON format.

After executing an action, \agent reflects on its effect by observing any state changes that may have occurred. During this stage, the agent is provided with available action types, general operating guidelines, the actual action taken, and its reasoning, as well as before-and-after UI states, represented by UI element representations and screenshots with SoM annotations. We request the agent to provide a concise summary of this step, including the intended action, success or failure, potential reasons for failure, and recommendations for subsequent actions. This summary will serve as the action history and be used for future action selection. See Appendix~\ref{sec:agent_details} for more details on the agent.

In addition to the full agent, we develop \agentsimple to measure the performance that can be achieved with minimal prompting, without guidelines or reflection mechanisms. This helps quantify the impact of more advanced prompting techniques and domain-specific guidance.

\subsubsection{SeeAct baseline}

We implement a baseline agent based on SeeAct~\citep{zheng2023seeact}, which was originally designed for GPT-4V for web navigation. Specifically, we implement the best-performing variant, SeeAct\textsubscript{choice}, which grounds actions via textual choices. We implement SeeAct for the Android environment to evaluate how an existing model that performs well on web tasks \citep{deng2023mind2web} can be adapted and applied to Android.

To accommodate the Android environment, we adapt SeeAct in several ways. Firstly, we augment the action space from the original SeeAct implementation to support actions needed for mobile, including scroll, long press, navigate home and back, and open app actions. Secondly, in lieu of the DOM, which is not available for Android apps, we utilize the accessibility tree to construct candidate UI actions. Due to the lack of the DOM representation, we do not use the bespoke ranker model from the original implementation. However, we observe that after applying a filtering heuristic to remove non-interactable elements, the majority of screens contains less than 50 candidate elements. See Appendix~\ref{sec:seeact_prompt} for more details on the implementation.

\begin{table}[t!]
\centering
\caption{Success Rates (SR) on \system and \mobileminiwob.}
\scalebox{0.84}{
\begin{tabular}{lllcc}
\toprule
Agent & Input & Base model & SR\textsubscript{\system} & SR\textsubscript{\mobileminiwob} \\ 
\midrule
\emph{Human} & \emph{screen}  & \emph{N/A} & \emph{\humanresult}  & \emph{100.0~~ }\\
\midrule
SeeAct~\citep{zheng2023seeact}  & SoM (screen + a11y tree)  & GPT-4 Turbo         & 15.5         & 66.1     \\
\midrule
M3A-Simple & a11y tree & Gemma 2 & ~~3.4 & 35.5 \\
M3A-Simple & a11y tree & Gemini 1.5 Pro & 14.7 & 55.2 \\
M3A-Simple & a11y tree & GPT-4 Turbo & 19.8 & \textbf{67.7} \\
\midrule
M3A & a11y tree & Gemma 2 & ~~9.5 & 45.6 \\
M3A & a11y tree & Gemini 1.5 Pro   & 19.4                       & 57.4       \\
M3A & SoM (screen + a11y tree)       & Gemini 1.5 Pro & 22.8                    & 40.3        \\
M3A & a11y tree & GPT-4 Turbo   & \textbf{30.6}                       & 59.7                 \\
M3A  & SoM (screen + a11y tree)       & GPT-4 Turbo         & 25.4                      & \textbf{67.7}\\
\bottomrule
\end{tabular}
}
\label{tab:agent_success_rates}
\end{table}

\subsection{Experimental results}
\label{sec:experiment_results}

We evaluate \agent, \agentsimple, and SeeAct on \system and \mobileminiwob. We set the seed to 30 and the temperature to 0 to aid reproducibility. Each task has a maximum allowed number of steps (detailed in Appendix~\ref{sec:task_list}), typically set to twice the number of steps needed by human annotators to complete the task. We use Gemini 1.5 Pro, GPT-4 Turbo, and the open-source Gemma 2 27B~\citep{gemmateam2024gemma2improvingopen} as base models. For \mobileminiwob, we evaluate on a subset of 62 tasks, consistent with recent studies \citep{zheng2024synapse, kim2024language, gur2022understanding}.

Table~\ref{tab:agent_success_rates} presents the success rates (SR) for the agents and human performance on both task suites. Although the agents have far from human performance, they demonstrate out-of-the-box capabilities in operating mobile UIs, exhibiting basic understanding and control capabilities of UIs. They can perform a variety of actions, including long-press, scrolling to search for information, and revising their plan if actions do not work out. The best performance is obtained by M3A when using GPT-4. On \system the SoM-based variant is less performant, while on \mobileminiwob it performs best. A similar result was obtained in recent work on computer agents for desktop applications~\citep{xie2024osworld}. We posit SoM plays a more critical role in \mobileminiwob tasks due to the often incomplete accessibility tree, compared to that of native Android apps.

The simplified agent variant \agentsimple shows a significant performance drop on \system tasks (19.8\% vs 30.6\% with GPT-4), indicating that additional prompting techniques and domain-specific guidance are beneficial for navigating the complexity of Android interactions. However, on \mobileminiwob tasks, \agentsimple achieves comparable performance (67.7\%), suggesting that these simpler tasks may not benefit as much from sophisticated prompting strategies. The open-source Gemma model's lower performance (9.5\% on \system, 45.6\% on \mobileminiwob) compared to proprietary models likely stems from its smaller parameter count, though exact comparisons are difficult as the parameter counts for GPT-4 and Gemini are not public.

\subsection{Analysis}
\label{sec:experiment_analysis}

Agents have difficulty understanding mobile UIs, often failing to detect visual cues that are essential for task completion (see Figure \ref{fig:perceptual_error}). Additionally, agents struggle with certain UI patterns and affordances, and when they make reasoning mistakes (see Figure \ref{fig:reasoning_error}), they often lack the capability to explore and adapt as humans do (see Figure \ref{fig:missing_knowledge}). Moreover, agents sometimes struggle with tasks that simply involve confirming system states, e.g., confirming the WiFi is turned on, suggesting challenges in both task and screen understanding.

The agents struggle with grounding, particularly when executing precise interactions, such as manipulating text (see Figure \ref{fig:grounding_errors}) or operating sliders, and they are often unable to recover from mistyping errors. In addition, for tasks that demand memory, such as performing transcriptions across apps, multiplying numbers, or scrolling, the agents struggle as they are unable to ``remember" content.

SeeAct performs less effectively than \agent on the \system task suite and similarly on \mobileminiwob, reflecting its optimization for web rather than mobile environments. It struggles with mobile-specific actions like long-presses and swipes, and often fails to select appropriate actions due to not incorporating screen elements during action generation. Memory-intensive tasks are particularly challenging, as SeeAct only caches actions without remembering outcomes, leading to repetitive, ineffective behaviors such as endless scrolling. This lack of quick error recovery often results in task termination once maximum steps are reached.

Finally, we note that large foundation models significantly increase latency, taking three times longer than humans on average to complete tasks. On average, \agent takes 3.9 minutes to complete a task, with the text-only version taking 2.5 minutes.

\section{Robustness Analysis}
\label{sec:robustness}

To understand agent robustness, we analyze M3A's performance across different random seeds, which generate different task parameters (e.g., calendar appointments, expense categories) and can consequently require different UI interaction patterns (e.g., scrolling to access hidden elements, handling varying numbers of elements to modify, or adapting to different input types and lengths). Across three seeds, we observe significant performance variations: 27.6\%, 26.3\% and 33.2\% (mean 29.0\%), obtained using M3A with GPT-4 Turbo with accessibility trees as input. Note that for consistency with existing literature we maintain the single-seed results in Table~\ref{tab:agent_success_rates}.

To better understand the sources of this variability, we evaluate agent robustness under two conditions: (1) identical tasks with the same parameters and (2) tasks with different parameter combinations, which change the initial state and task definition. We perform this analysis on a representative subset of \system tasks that span different interaction patterns and complexity levels (listed in Appendix~\ref{sec:agent_robustness_appendix}). Due to computational constraints, we conduct 20 trials for each task using our strongest agent configuration - \agent using the accessibility tree and GPT-4. 

\begin{figure}[t]
\centering
\includegraphics[width=0.7\textwidth]{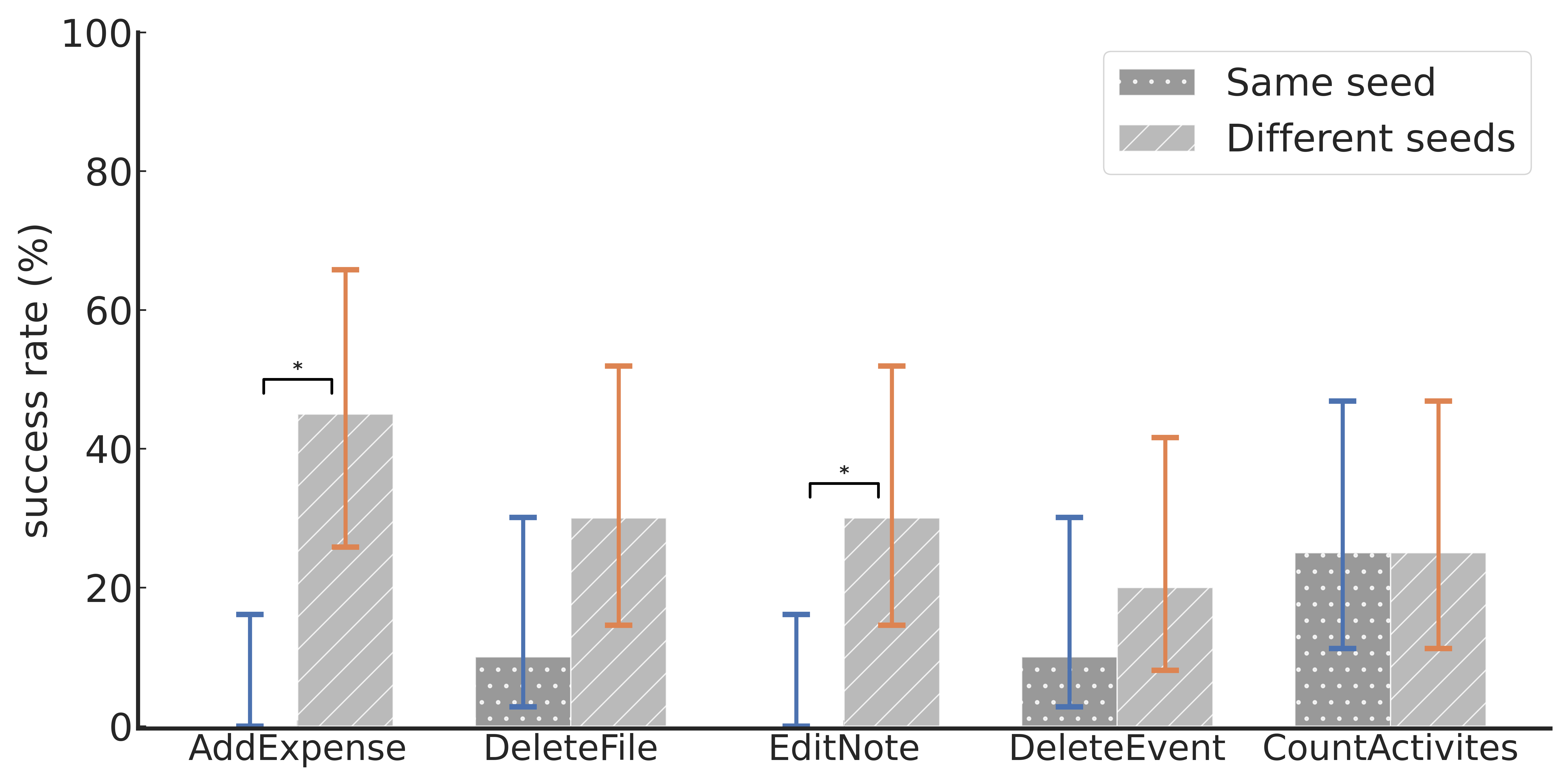}
\caption{Success rate variation across tasks due to the parametrization built into \system. Using a fixed seed, the agent appears completely incapable of solving some tasks due to ``bad luck" with the seed. In contrast, under different task parameterizations, we observe the agent can solve the tasks fairly often. Wilson binomial proportion confidence intervals (95\%) are shown for the different seed group (orange) and the same seed group (blue). The different seed group has higher variance than the same seed group. Significant differences, with p-value \textless~ 0.05, are indicated by ``*".}
\vspace{-1ex}
\label{fig:robustness}
\end{figure}

Figure \ref{fig:robustness} shows our results. With a constant seed, the agent fails on add and edit tasks and rarely solves delete tasks, primarily due to UI operation challenges. Surprisingly, performance varies even with a fixed seed, suggesting model non-determinism affects reliability. Performance varies significantly more with different seeds, with statistically significant differences for add expense and edit note tasks. The high intra-task variation indicates the model's sensitivity to task parameters. Section~\ref{sec:agent-struggles} provides an analysis on how specific parameter variations impact agent performance.

This sensitivity aligns with observations in RL research \citep{henderson2018deep, raffin2021stable, colas2018many}, suggesting performance is best represented by the mean across seeds. We believe \system's support for such analysis will become increasingly valuable as more efficient models are developed. Finally, we note the observation of non-zero rewards under some seeds points to potential enhancements through RL-like mechanisms in future work. 

To assess AndroidWorld’s robustness to OS variations, we tested on a Pixel 5 (Android 12) alongside our primary setup (Pixel 6, Android 13). The agent achieved a 28.4\% success rate, with performance variations akin to those from random seed changes, suggesting it maintained its capabilities despite differing UI layouts and device types.

These experiments underscore the importance of testing agents under varied conditions, a capability that \system effectively supports.

\section{Conclusion}

We introduced \system, a realistic and robust agent environment for Android that enables the development and evaluation of autonomous agents across a wide range of tasks and apps. \system provides a reproducible task suite consisting of \ntasks tasks across \napps apps, with each task dynamically generated using random parameters to challenge agents with millions of unique goals. By releasing \system and establishing benchmark performance with \agent, we aim to accelerate research and development in this area, ultimately leading to the creation of computer use agents capable of operating effectively in real-world environments. Further, the dynamic nature of \system opens up new research opportunities for online learning algorithms in computer use agents. 

\bibliographystyle{iclr2025_conference}
\bibliography{references}

\begin{appendices}

\section{Limitations}
\label{sec:limitations}
\system currently supports open-source Android apps (\textgreater1M downloads) and built-in system apps. While testing on trending apps would be desirable, we found open-source apps often present harder challenges due to their less-optimized UIs. Popular apps typically offer more shortcuts and UI affordances, while open-source apps may require more complex interaction patterns. For example, in Figure \ref{fig:missing_knowledge}, the agent fails by repeatedly searching for a non-existent "delete-all" button instead of recognizing the need to delete notes individually.

\section{Ethical considerations}
\label{sec:ethical_considerations}

\paragraph{Malicious use} There is a risk that malicious actors could engineer agents to bypass security measures like CAPTCHAs or engage in activities like spamming. Additionally, they could alter prompts or screen outputs to further harmful objectives.

\paragraph{Societal impact} Automation agents may transform societal norms, disrupt employment, and modify human behavior. While they can enhance efficiency, this improvement could pose risks if exploited by malevolent forces.

\section{\system environment}
\label{sec:appendix_environment}

\subsection{Observation Space}
\label{subsec:observation}

In \system, the Android screen is represented using a \texttt{State} class, which includes the following attributes:
\begin{itemize}
\item \textbf{Pixels}: An RGB array representing the current screen capture of the device. The screenshot resolution is $2400 \times 1080 \times 3$.
\item \textbf{Accessibility tree}: A raw representation of the accessibility tree.\footnote{Represented using all current windows; \url{https://developer.android.com/reference/android/view/accessibility/AccessibilityWindowInfo}} This UI tree provides a detailed snapshot of all UI elements currently displayed on the screen. We utilize an accessibility forwarding app from AndroidEnv \citep{android_env}, which leverages gRPC to transmit the accessibility tree data efficiently to the device.
\item \textbf{UI elements}: A list of processed UI elements extracted from the children of the accessibility tree. Each \texttt{UIElement} contains attributes such as text, content description, bounding boxes, and various state flags (e.g., clickable, scrollable, focused).
\end{itemize}

Since Android observations and actions are asynchronous, changes resulting from actions may take some time to manifest. Therefore, instead of using an RL-based interface, which assumes a tight coupling between actions and observations, we design an interface for the agent tailored for asynchronous interaction. This interface implements a \texttt{get\_state} method responsible for capturing the current state of the environment, typically after executing an action. This method includes an optional \texttt{wait\_to\_stabilize} flag, which, when enabled, employs heuristics to ensure the UI elements are not in a transient state, thus providing a stable and accurate snapshot of the environment.

\subsection{Action space}

Actions are stored using a Python dataclass and executed using \texttt{adb}. The action space includes:

\begin{itemize}
\item Direct UI Actions:
    \begin{itemize}
    \item Click-based actions (click, long press): Simulates touch events at specified coordinates
    \item Text input: Simulates typing in focused text fields
    \item Navigation: Sends home/back key events
    \item Scrolling: Executes swipes in four directions (up, down, left, right)
    \item App launching: Starts specified applications
    \end{itemize}

\item Task Management Actions:
    \begin{itemize}
    \item Status: Reports if task is in-progress, complete, or infeasible
    \item Answer: Provides responses, which are needed for information retrieval tasks
    \end{itemize}

\item System Actions:
    \begin{itemize}
    \item Wait: No-op useful for loading screens and UI transitions
    \item Unknown: No-op for handling internal errors
    \end{itemize}
\end{itemize}

\begin{lstlisting}[language=Python, caption={Pseudo-code representation of the action space.}, basicstyle=\ttfamily\footnotesize, style=mystyle]
ACTION_TYPES = {
    # UI Manipulation
    "CLICK": "click",
    "SCROLL": "scroll",
    "INPUT_TEXT": "input_text",
    "NAVIGATE_HOME": "navigate_home",
    "NAVIGATE_BACK": "navigate_back",
    "KEYBOARD_ENTER": "keyboard_enter",
    "OPEN_APP": "open_app",
    "LONG_PRESS": "long_press",
    
    # Control Flow
    "STATUS": "status",    # Reports task completion state
    "WAIT": "wait",       # Handles UI transitions
    "ANSWER": "answer",   # For information retrieval tasks
    "UNKNOWN": "unknown"  # No-op for internal errors
}

@dataclasses.dataclass()
class JSONAction:
  """Represents a parsed JSON action.

  # Example
  result_json = {'action_type': 'click', 'x': %d, 'y': %d}
  action = JSONAction(**result_json)

  Attributes:
    action_type: The action type.
    index: The index to click, if action is a click. Either an index or a <x, y>
      should be provided. See x, y attributes below.
    x: The x position to click, if the action is a click.
    y: The y position to click, if the action is a click.
    text: The text to type, if action is type.
    direction: The direction to scroll, if action is scroll.
    goal_status: If the status is a 'status' type, indicates the status of the goal.
    app_name: The app name to launch, if the action type is 'open_app'.
  """
  action_type: str
  index: int = None
  x: int = None
  y: int = None
  text: str = None
  direction: str = None
  goal_status: str = None
  app_name: str = None
\end{lstlisting}

In addition to the UI-based action space described above, AndroidWorld provides a set of high-level APIs for direct device interaction (i.e., sending SMS messages, opening web pages, managing contacts). While the core action space focuses on fundamental UI control capabilities, these supplementary APIs found in \texttt{env/tools.py} enable future research into hybrid interaction approaches that combine both UI-based and programmatic device control.

\subsection{\mobileminiwob}
\label{sec:miniwob_details}

\begin{figure}[t]
  \centering
  \includegraphics[width=0.22\linewidth]{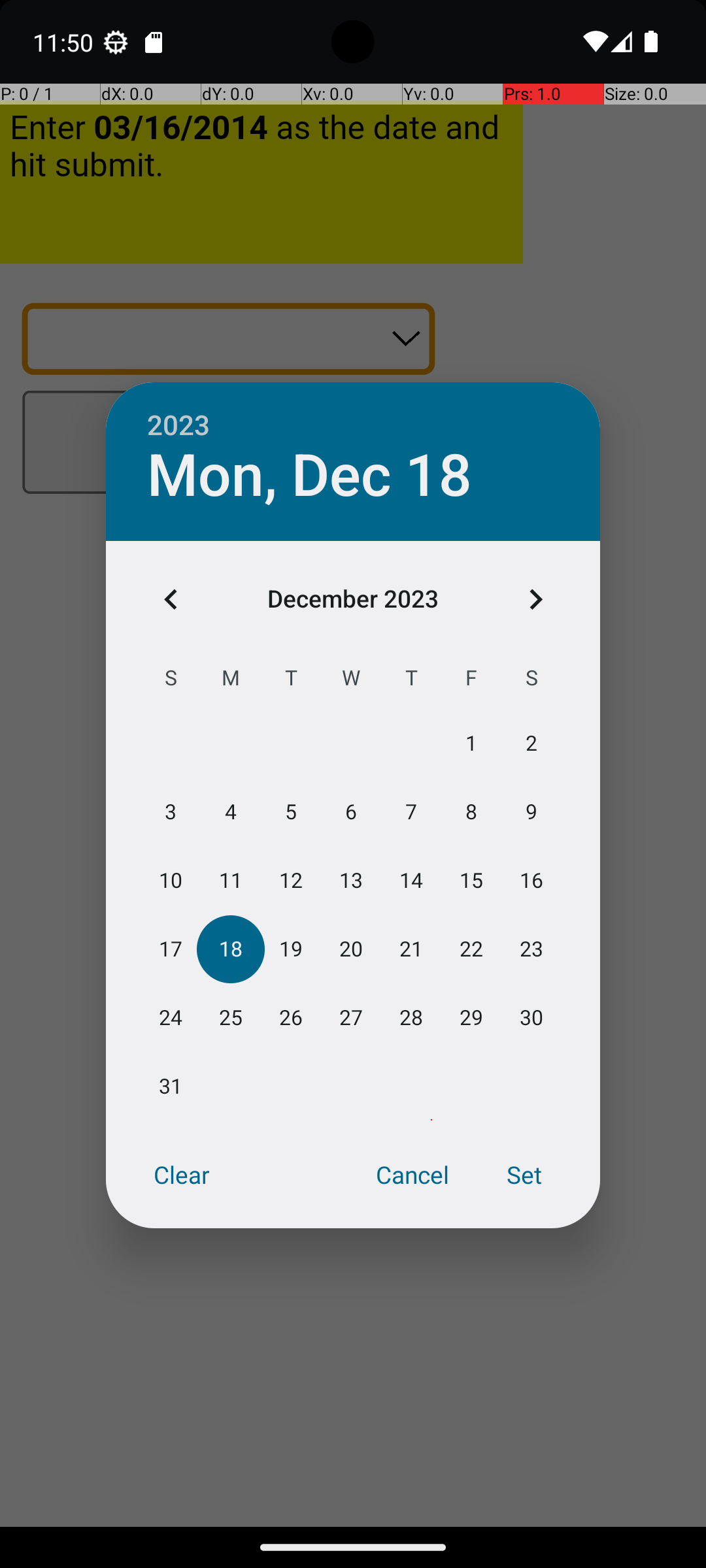}
  \caption{Native Android UI widget rendering for HTML5 \lstinline|<input>| element.}
  \label{fig:miniwob_android_widget}
\end{figure}

Authors manually completed all tasks in \mobileminiwob, implemented as a WebView app, to verify solvability on a mobile interface. \mobileminiwob differs from \miniwob due to the touch-based interface, which required different approaches for certain tasks. For instance, highlighting text from the \texttt{highlight-text} tasks involves using Android's long-press and cursor-moving functionalities. HTML5 \lstinline|<input>| elements are natively rendered with native Android UI widgets like the date-picker (see Figure~\ref{fig:miniwob_android_widget}).

Our implementation of \miniwob contains 92 tasks in total. We exclude the following tasks: \texttt{chase-circle} (requires near-realtime movement, unachievable by humans on emulators), \texttt{moving-items} (too hard to click in emulator), \texttt{drag-cube} (drags will scroll the screen, moving the task out of view), \texttt{drag-items-grid} (elements are not interactable on Android), \texttt{drag-items} (elements are not interactable on Android), \texttt{drag-shapes} (drags will scroll the screen, moving the task out of view), \texttt{drag-sort-numbers} (elements are not interactable on Android), \texttt{text-editor} (cannot underline everything, weird glitch), \texttt{number-checkboxes} (not correctly rendered: only three columns), \texttt{use-slider-2} (slider implementation not working), \texttt{use-spinner} (slider implementation not working), and \texttt{click-menu} (the menu responsiveness breaks and the task does not behave as intended).

\section{\system Benchmark Details}
\label{sec:benchmark_details}

\subsection{App selection}

Our selection of apps (summarized in Table~\ref{table:app-list}) was guided by three main factors: use case, popularity, and the need for consistency and reproducibility.

\paragraph{Use case and categories} We analyzed popular app categories in app stores, focusing on productivity, communication, and multimedia. Selected apps had to meet criteria such as not requiring a login and storing data locally on the device. Additionally, we considered apps from categories that the authors commonly used, ensuring the selection was representative of real-world Android usage.

\paragraph{Popularity} We used download statistics from the Google Play Store to gauge app popularity, selecting apps with over 1 million downloads. Most of the selected apps exceeded this threshold. Less popular apps were also included if they featured common UI patterns and affordances, ensuring they are indicative of typical Android app usage. For instance, Simple Calendar Pro, though less downloaded, has a UI comparable to the widely-used Google Calendar app.

\paragraph{Consistency and reproducibility} All apps were sourced from F-Droid, an open-source Android app repository. This allowed us to manage app versions precisely by selecting and distributing specific APKs. We use the newest version of each app at the time of download.

\begin{table*}
\centering
\caption{List of \system apps and number of tasks for each one.}
\scalebox{0.88}{
    \begin{tabular}{lp{10cm}c}
    \toprule
    \textbf{App name} & \textbf{Description} & \textbf{\# tasks} \\
    \midrule
    Simple Calendar Pro & A calendar app for creating, deleting, and managing events and appointments. & 17~~ \\

    Settings & The Android system settings app for managing device settings such as Bluetooth, Wi-Fi, and brightness. & 15~~ \\
   
    Markor & A note-taking app for creating, editing, deleting, and managing notes and folders. & 14~~ \\

    Broccoli - Recipe App & A recipe management app for adding, deleting, and organizing recipes. & 13~~ \\
  
    Pro Expense & An expense tracking app for adding, deleting, and managing expenses. & 9 \\
  
    Simple SMS Messenger & An SMS app for sending, replying to, and resending text messages. & 7 \\
  
    OpenTracks & A sport tracking app for recording and analyzing activities, durations, and distances. & 6 \\

    Tasks & A task management app for tracking tasks, due dates, and priorities. & 6 \\

    Clock & An app with stopwatch and timer functionality. & 4 \\

    Joplin & A note-taking app. & 4 \\

    Retro Music & A music player app. & 4 \\
 
    Simple Gallery Pro & An app for viewing images. & 4 \\

    Camera & An app for taking photos and videos. & 3 \\
 
    Chrome & A web browser app. & 3 \\

    Contacts & An app for managing contact information. & 3 \\

    OsmAnd & A maps and navigation app with support for adding location markers, favorites, and saving tracks. & 3 \\

    VLC & A media player app for playing media files. & 3 \\

    Audio Recorder & An app for recording and saving audio clips. & 2 \\
  
    Files & A file manager app for the Android filesystem, used for deleting and moving files. & 2 \\
  
    Simple Draw Pro & A drawing app for creating and saving drawings. & 1 \\
    \bottomrule
    \end{tabular}
}
\label{table:app-list}
\end{table*}

\subsection{Task classification and generation}
\label{sec:task_definition}

We categorize tasks into two types: those with side-effects and those without. Tasks with side-effects are those that modify the internal state of the device or applications, such as turning off Wi-Fi or creating a calendar event. These tasks are implemented as distinct Python classes, each with its own parameter generation, initialization, evaluation, and teardown methods. 

Below we show an example of the task evaluation for a \texttt{SendSms} task, which involves sending and validating a text message. The pseudocode illustrates the task initialization, success check, and parameter generation methods. Each task has its own random parameter generation method and success logic.

\begin{lstlisting}[language=Python, basicstyle=\ttfamily\footnotesize, style=mystyle]
class SendSms(TaskEval):
  """Task sending and validating a text message has been sent.

  It checks the SMS telephony database, which is located at:
  /data/data/com.android.providers.telephony/databases/mmssms.db."""

  template = (
      "Send a text message using Simple SMS Messenger to "
      "{number} with message: {message}"
  )

  def initialize_task(self, env: interface.AsyncEnv) -> None:
    """Sets up the initial state of the task."""
    super().initialize_task(env)
    clear_sms_database(env.base_env)

  def is_successful(self, env: interface.AsyncEnv) -> float:
    """Checks if the SMS was sent successfully."""
    super().is_successful(env)
    messages = get_messages(env.base_env)
    return check_message_exists(
        phone_number=self.params["number"],
        body=self.params["message"],
    )

  def teardown(self, env: interface.AsyncEnv) -> None:
    """Clears the SMS database."""
    super().teardown(env)
    clear_sms_database(env.base_env)

  @classmethod
  def generate_random_params(cls) -> dict[str, Any]:
    number = generate_random_number()
    message = generate_random_message()
    return {
        "number": number,
        "message": message,
    }
\end{lstlisting}

\subsection{Information retrieval tasks}
\label{sec:info_retrieval_details}

Tasks without side-effects are Information Retrieval tasks, requiring the agent to answer a question based on the device or app's current state. For these tasks, instead of a Python class, we create a protobuf structure to specify the prompt, parameter values, and initialization and validation logic. We decided to use a structured data format with the belief that it would allow us to define new information retrieval tasks by simply adding new entries, making it easier to scale up the number of tasks without needing to write and maintain Python classes for each one.

Initialization is defined per app, including only the state relevant to the prompt's answer and exclusion conditions for generating random states. This ensures that no random state contains information that could alter the expected answer. The initial state and prompt are parameterized using random values from the specified task parameters. For validation, we define the expected answer format within the prompt and use a few supported functions (``count", ``sum", ``identity") to generate the answer from the initial state.

Once an app and its specific logic are programmed, new tasks can be generated using an LLM to generate the task's protobuf. The process is not automatic and requires human review. Common issues with LLM-generated tasks include missing fields, hallucinated fields, incompatible parameter generation, insufficient parameter usage, and non-specific task prompts. We observed that the complexity of the proto structure correlates with an increase in generated task issues. Despite these challenges, we found that editing LLM-generated protobufs can be more efficient than writing a complete task from scratch.

Below we show a simplified version of the task definition for the \texttt{SimpleCalendarEventsOnDate} task which involves checking which events are on a certain date. It specifies the relevant event, the exclusion conditions for any noisy event, how to determine success, and possible parameter values to be chosen at random that will be used to fill out the task definition.

\begin{lstlisting}[language=Java, basicstyle=\ttfamily\footnotesize, style=mystyle]
tasks {
  name: "SimpleCalendarEventsOnDate"
  prompt: "What events do I have {date} in Simple Calendar Pro? Answer with the titles only. If there are multiple titles, format your answer as a comma separated list."
  complexity: 1
  relevant_state {
    // Defines information for the goal events.
    state: {
      calendar {
        events {
          start_date: "{date}"
          start_time: "{time}"
          duration: "{duration}"
          title: "{title}"
        }
      }
    }
    // Non-goal events.
    exclusion_conditions {
      field: "start_date"
      operation: EQUAL_TO
      value: "{date}"
    }
  }
  success_criteria {
    expectations {
      field_transformation {
        operation: IDENTITY
        field_name: "title"
      }
      match_type: STRING_MATCH
    }
  }

  task_params {
    name: "time"
    possible_values: "11:00am"
    // ...
  }

  task_params {
    name: "date"
    possible_values: "October 15 2023"
    //...
  }
  task_params {
    name: "duration"
    possible_values: "30 m"
    // ...
  }
  task_params {
    name: "title"
    possible_values: "Data Dive"
    // ...
  }
}
\end{lstlisting}

\subsection{Humans for task analysis}
\label{sec:human_performance}

During development, we recruited six volunteers with proficient programming skills to analyze task difficulty, duration, and category. Each human was assigned an equal portion of tasks and tasked with identifying bugs during this annotation phase. This process resulted in the discovery and resolution of over 30 bugs.

To evaluate human performance, we enlisted two software engineers to complete the tasks using an Android emulator. Participants were provided with task descriptions and attempted to achieve the goals based on their interpretations. Each participant had one attempt per task. The majority of errors stemmed from misinterpretations or minor errors, such as entering an incorrect file extension. Other errors occurred when participants encountered unfamiliar user interfaces, impeding their ability to solve the tasks on their first attempt.

In both exercises, we informed participants about the intended use of the collected data. Participants were not required to enter any personal information in the tested tasks. 

\subsection{Task examples}

Table~\ref{table:task-list} lists some additional examples of tasks and highlights which task attributes can be parameterized in unlimited ways. 

\begin{table*}[h]
\centering
\caption{Examples of \system tasks. We list the task nickname, the task template indicating which task attributes can be parameterized, the initialization logic that is executed before the task starts and pseudo code describing the success evaluation.} 
\scalebox{0.68}{
    \begin{tabular}{p{3.3cm}p{5.9cm}p{3.6cm}p{5.4cm}}
    \toprule
    \textbf{Task nickname} & \textbf{Task template} & \textbf{Initialization logic} & \textbf{Success evaluation code} \\
    \midrule
    VlcCreatePlaylist & Create a playlist in VLC, titled ``\{playlist\_name\}" with the following files, in order: \{files\} & Create new mpeg files: files + ``noise" files that should not be added. Add them to VLC videos folder. & \texttt{execute\_sql(vlc\_query) == files} \\
     \midrule
    RecipeAddMultiple RecipesFromImage & Add the recipes from recipes.jpg in Simple Gallery Pro to the recipe app. & Write a receipt file with recipes to Simple Gallery. & \texttt{sql\_rows\_exist(expected\_recipes)} \\
    \midrule
    MarkorEditNote & Edit \{file\_name\} in Markor. \{file\_operation\}. & Generate file with starting content, along with ``noise" files not relevant to goal. \textit{Note:} \texttt{file\_operation} \textit{can be to add a footer, header, or update note content.}  & \texttt{file\_exists(file\_name, content=expected\_content)} \\
    \midrule
    ExpenseAddSingle & Add the following expenses into pro expense: \{expense\_csv\} & Add to the app's SQLite database the expense that should be deleted, along with ``noise" expenses that should not be deleted. & \texttt{sql\_rows\_exist(expense\_obj)} \\
    \midrule
    SimpleCalendarDelete EventsOnRelativeDay & In Simple Calendar Pro, delete all events scheduled for this \{day\_of\_week\}. & add to the app's SQLite database calendar events on specified day, along with ``noise" events that should not be deleted. &  \texttt{!sql\_rows\_exist(expected\_events)} \\
    \midrule
    FilesDeleteFile & Delete the file \{file\_name\} from the Android filesystem located in the \{subfolder\} folder within the sdk\_gphone\_x86\_64 storage area. & Generate specified file, along with ``noise" files that should not be deleted. & \texttt{!file\_exists(file\_name)}\\
    \midrule
    SportsTrackerActivities CountForWeek & How many \{category\} activities did I do this week in the OpenTracks app? Express your answer as a single integer. & add to the app's SQLite database activities for the specified category, along with ``noise" activities.  & \texttt{int(agent\_response) == expected\_count} \\
    \bottomrule
    \end{tabular}
}
\label{table:task-list}
\end{table*}

\section{\system agent details}
\label{sec:agent_details}

\subsection{\agent observations}

\system consumes the raw screen pixels, the screen shot with Set-of-Mark (SoM)~\citep{yang2023set} annotations, and a list of UI elements on screen.

\begin{lstlisting}[basicstyle=\ttfamily\footnotesize, style=mystyle, caption={ The prompt format pertaining to screen representation with UI elements.}]
Here is a list of descriptions for some UI elements on the current screen:

UIelement0: UIElement(text="VLC", content_description=None, class_name="android.widget.EditText",
bbox_pixels=BoundingBox(x_min=98, x_max=886, y_min=146, y_max=311), ...)
UIelement1: UIElement(text=None, content_description="Clear search box", class_name="android.widget.ImageButton",
bbox_pixels=BoundingBox(x_min=886, x_max=1023, y_min=160, y_max=297), ...)
UIelement2: UIElement(text="15:11", content_description="15:11", class_name="android.widget.TextView",
bbox_pixels=BoundingBox(x_min=50, x_max=148, y_min=1, y_max=128), ...)
... More elements listed ...

... Guidelines on action selection emitted ...

Now output an action from the above list in the correct JSON format, following the reason why you do that. Your answer should look like:

Reason: ...
Action: {"action_type":...}
\end{lstlisting}

\subsection{\agent actions}

\begin{figure}[h]
  \centering
  \includegraphics[scale=0.1]{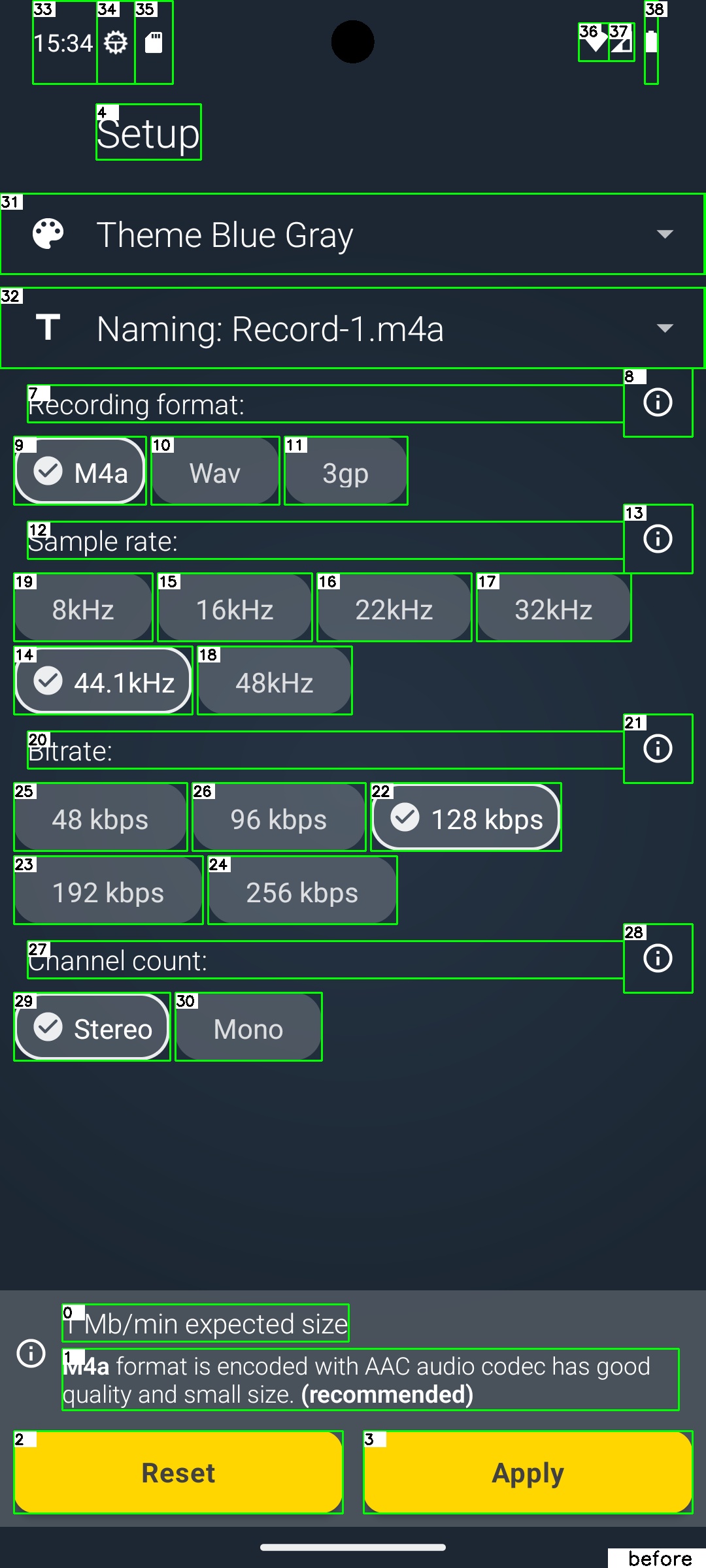}
  \caption{Set-of-marks overlaid on an Android screen.}
  \label{fig:som_example}
\end{figure}

For the SoM prompting, the screen is annotated based on the UI elements extracted from the accessibility tree, which form the agent's action space. Figure~\ref{fig:som_example} shows one example. 

\subsection{Error analysis}
\label{sec:error_analyses}

We analyze \agent errors based on broader categories we observe during evaluation.

\paragraph{Perceptual errors}
Perceptual errors are caused when the model fails to recognize crucial elements on the screen necessary for task completion.

\begin{figure*}[t!]
  \centering
  \begin{subfigure}[t]{0.325\linewidth}
  \centering
    \includegraphics[width=0.9\linewidth]{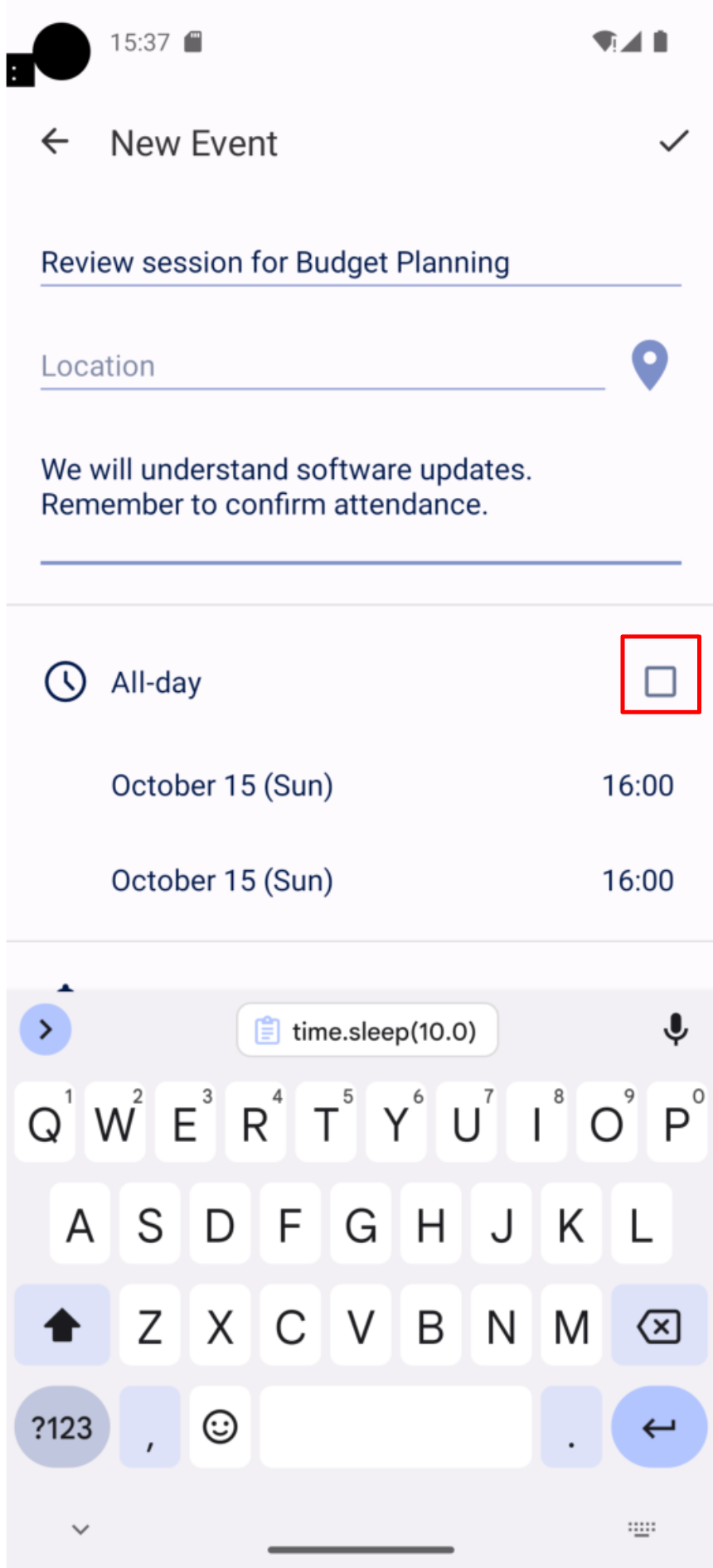}
   \caption{Perceptual error. Red square highlights issue.}
   \label{fig:perceptual_error}
\end{subfigure}
\hfill
\begin{subfigure}[t]{0.325\linewidth}
\centering
    \includegraphics[width=0.9\linewidth]{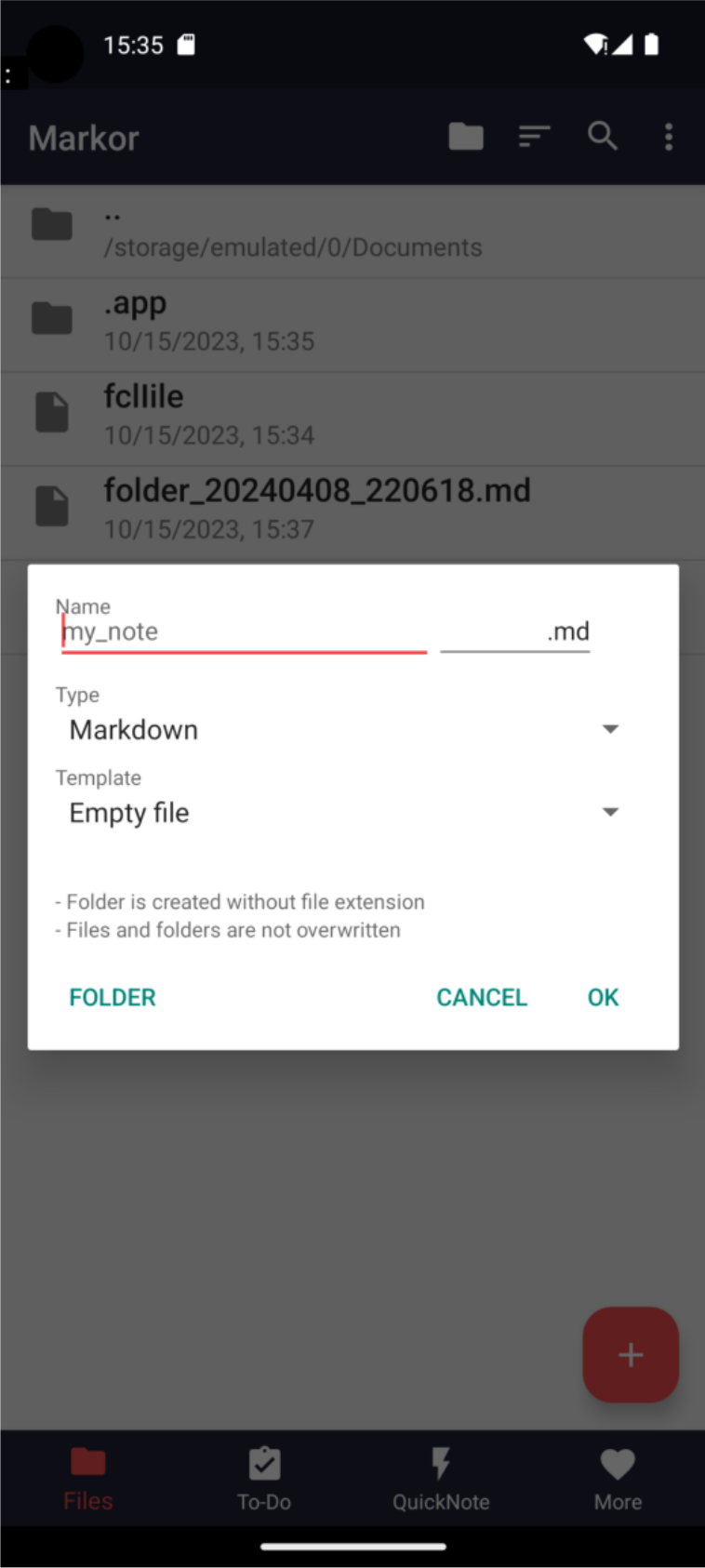}
  \caption{Reasoning error. The agent's next action is to start entering the note's contents, which is incorrect because it needs to enter the note's name first.}
  \label{fig:reasoning_error}
\end{subfigure}
\hfill
\begin{subfigure}[t]{0.325\linewidth}
\centering
    \includegraphics[width=0.9\linewidth]{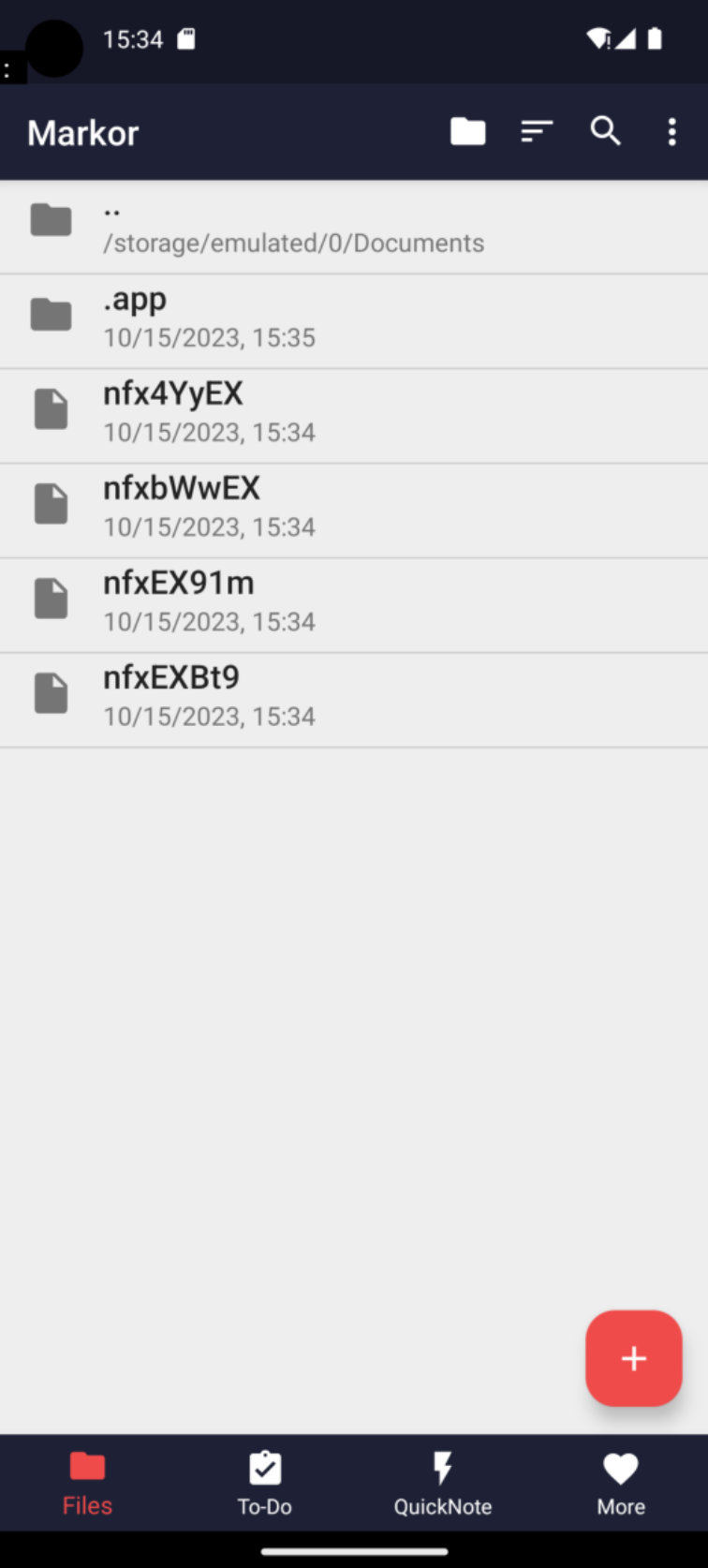}
  \caption{Missing knowledge error. To delete all notes, the agent mistakenly looks for an option to delete all the notes at once, rather than trying to do it note-by-note.}
  \label{fig:missing_knowledge}
\end{subfigure}
\caption{Screens illustrating perceptual, reasoning and missing knowledge errors.}
\label{fig:perceptual_and_reasoning_errors}
\end{figure*}

For the task below, the model does not recognize that the “All-day” checkbox is currently not checked (see Figure~\ref{fig:perceptual_error}):

\begin{quote}
\emph{In Simple Calendar Pro, create a recurring calendar event titled 'Review session for Budget Planning' starting on 2023-10-15 at 14h. The event recurs weekly, forever, and lasts for 60 minutes each occurrence. The event description should be 'We will understand software updates. Remember to confirm attendance.'}
\end{quote}

\paragraph{Reasoning errors}
Reasoning errors occur when the model misinterprets the task requirements or the current state, leading to incorrect actions.

For example, in the task below, the model mistakenly believes the note name has already been entered, so it types the note text into the “Name” field and cannot recover from this error (see Figure~\ref{fig:reasoning_error}):

\begin{quote}
\emph{Create a new note in Markor named ipsum\_mNcKd36pDq.txt with the following text: ...}
\end{quote}

Here is the agent's reasoning trace: 

\begin{quote}
\emph{The user wants to create a new note with a specific name and text. The name of the note has already been entered, so all that needs to be done is to enter the text and click on the OK button.}
\end{quote}

\paragraph{Missing knowledge errors}
Missing knowledge errors occur when the model lacks the necessary understanding of the application's UI to complete a task efficiently.

For example, in the task below, the agent does not know how to delete all files at once. It looks for an option to do that instead of deleting each file one by one (see Figure~\ref{fig:missing_knowledge}):

\begin{quote}
\emph{Delete all files in the current directory.}
\end{quote}

\paragraph{Grounding errors}
Grounding errors occur when the model fails to correctly interact with the UI elements based on their spatial or contextual positioning.

\begin{figure*}[t]
  \centering
  \begin{subfigure}[t]{0.43\linewidth}
  \centering
    \includegraphics[width=0.6\linewidth]{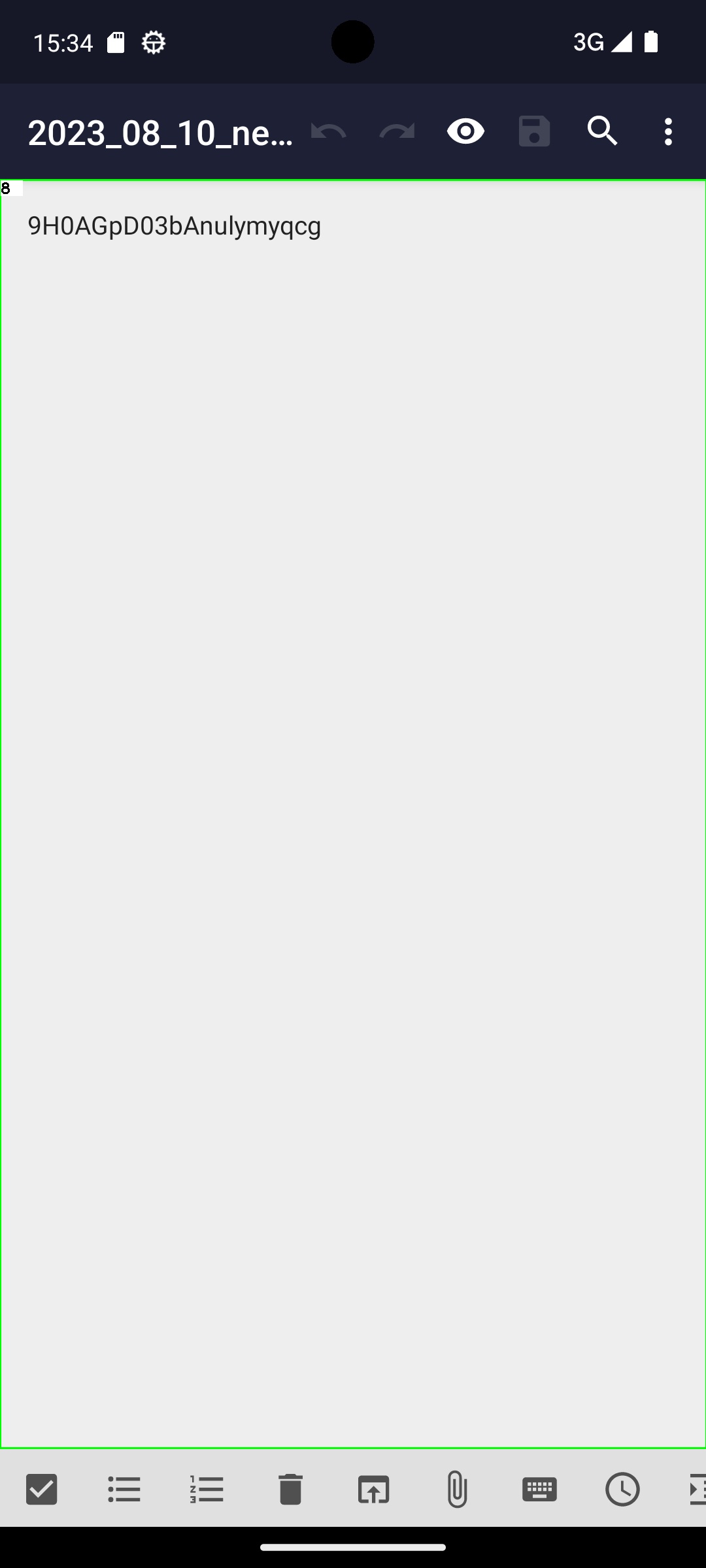}
   \caption{Error on initial click.}
   \label{fig:ground_error}
\end{subfigure}
\begin{subfigure}[t]{0.45\linewidth}
\centering
    \includegraphics[width=0.57\linewidth]{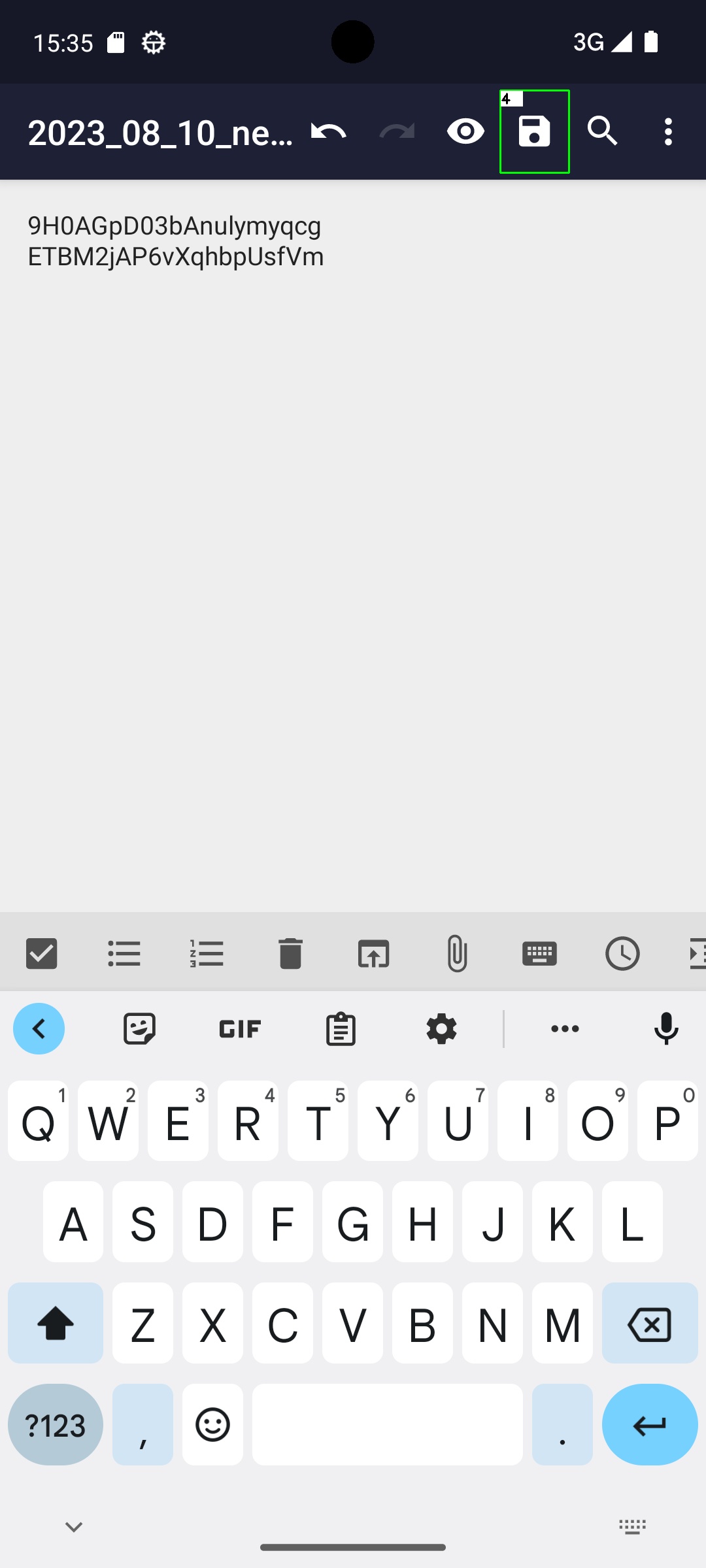}
  \caption{Error with text entered and saving.}
  \label{fig:ground_error2}
\end{subfigure}
\caption{Screens illustrating grounding errors.}
\label{fig:grounding_errors}
\end{figure*}

For the task below, the agent needs to update the Markor note by prepending text to the existing text. Figure~\ref{fig:grounding_errors} illustrates the errors the agent makes. It clicks the entire text field area, highlighted in green, which automatically places the cursor after the current text, resulting in the new text being appended after the current content.

\begin{quote}
\emph{Update the Markor note `2023\_08\_10\_neat\_wolf.txt` by adding the following text, along with a new blank line before the existing content: "ETBM2jAP6vXqhbpUsfVm", and rename it to `sure\_ocean\_uRnI.txt`.}
\end{quote}

Then, in the next screen, the text has been entered after the existing content, and the agent clicks the save button.

\subsection{Agent robustness experiments}
\label{sec:agent_robustness_appendix}

We ran the agent on the following tasks (the nicknames shown in the figures in parentheses):

\begin{itemize}
  \item \texttt{MarkorEditNote} (EditNote)
  \item \texttt{ExpenseAddSingle} (AddExpense)
  \item \texttt{SimpleCalendarDeleteEventsOnRelativeDay} (DeleteEvent)
  \item \texttt{FilesDeleteFile} (DeleteFile)
  \item \texttt{SportsTrackerActivitiesCountForWeek} (CountActivities)
\end{itemize}

More details about these tasks can be found in Table~\ref{table:task-list}.

\subsection{Agent struggles due to task parameterization}
\label{sec:agent-struggles}

\begin{figure}[h]
  \centering
  \includegraphics[scale=0.35]{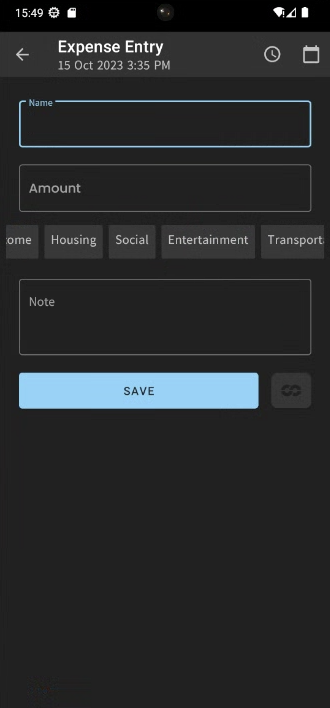}
  \caption{The expense entry interface features a horizontally scrollable category selector. When certain parameterization seeds require selecting categories that are not initially visible (e.g., ``Food""), the agent fails to discover the scrolling interaction required to access them.}
  \label{fig:horizontal_slider}
\end{figure}

The variance in success rates (Figure \ref{fig:robustness}) across different seeds demonstrates how task parameterization fundamentally changes task difficulty. For instance, in the \texttt{ExpenseAddSingle} task, the seed determines which expense category must be selected (see UI in Figure~\ref{fig:horizontal_slider}). When the seed specifies readily on-screen visible categories (e.g., "Housing", "Social"), the agent can complete the task. However, when the seed requires categories that are only accessible via horizontal scrolling (e.g., "Food", "Other"), the agent consistently fails due to its inability to discover and execute this UI interaction pattern. 

Similarly, the \texttt{MarkorEditNote} task's difficulty varies based on the seed-determined variant: adding text to the top of a note, adding text to the bottom, or replacing existing text. The ``replace" variant requires a more complex sequence of UI interactions (long-press, text selection, deletion, then text entry) compared to the simpler "header" variant. This explains both the complete failure under fixed seeds that happen to select challenging variants, and the higher but variable success rates when using different seeds that allow the agent to encounter various task parameterizations.

\subsection{SeeAct details}
\label{sec:seeact_prompt}

We modify the SeeAct prompt \citep{zheng2023seeact} to reflect that the environment is Android by inputting elements from the accessibility tree and supporting additional actions (e.g., scrolling). Below we include the updated prompt. We annotate the system, user, and assistant roles that are each provided to the OpenAI API.

\begin{lstlisting}[basicstyle=\ttfamily\footnotesize,breaklines=true,language={}, style=mystyle, keywordstyle=\color{black}]
  
> Role: SYSTEM
Imagine that you are imitating humans operating an Android device for a task step by step. At each stage, you can see the Android screen like humans by a screenshot and know the previous actions before the current step decided by yourself through recorded history. You need to decide on the first following action to take. You can tap on an element, long-press an element, swipe, input text, open an app, or use the keyboard enter, home, or back key. (For your understanding, they are like `adb shell input tap', `adb shell input swipe', `adb shell input text', `adb shell am start -n', and `adb shell input keyevent'). One next step means one operation within these actions. Unlike humans, for typing (e.g., in text areas, text boxes), you should try directly typing the input or selecting the choice, bypassing the need for an initial click. You should not attempt to create accounts, log in or do the final submission. Terminate when you deem the task complete or if it requires potentially harmful actions.

> Role: USER
You are asked to complete the following task: <GOAL>

Previous Actions:
<PREVIOUS ACTIONS>

The screenshot below shows the Android screen you see. Follow the following guidance to think step by step before outlining the next action step at the current stage:

(Current Screen Identification)
Firstly, think about what the current screen is.

(Previous Action Analysis)
Secondly, combined with the screenshot, analyze each step of the previous action history and their intention one by one. Particularly, pay more attention to the last step, which may be more related to what you should do now as the next step. Specifically, if the last action involved a INPUT TEXT, always evaluate whether it necessitates a confirmation step, because typically a single INPUT TEXT action does not make effect. (often, simply pressing 'Enter', assuming the default element involved in the last action, unless other clear elements are present for operation).

(Screenshot Details Analysis)
Closely examine the screenshot to check the status of every part of the screen to understand what you can operate with and what has been set or completed. You should closely examine the screenshot details to see what steps have been completed by previous actions even though you are given the textual previous actions. Because the textual history may not clearly and sufficiently record some effects of previous actions, you should closely evaluate the status of every part of the screen to understand what you have done.

(Next Action Based on Android screen and Analysis)
Then, based on your analysis, in conjunction with human phone operation habits and the logic of app design, decide on the following action. And clearly outline which element on the Android screen users will operate with as the first next target element, its detailed location, and the corresponding operation.

To be successful, it is important to follow the following rules:
1. You should only issue a valid action given the current observation.
2. You should only issue one action at a time
3. For handling the select dropdown elements on a screen, it's not necessary for you to provide completely accurate options right now. The full list of options for these elements will be supplied later.

> Role: ASSISTANT
<AGENT RESPONSE TO ABOVE>

> Role: USER
(Reiteration)
First, reiterate your next target element, its detailed location, and the corresponding operation.

(Multichoice Question)
Below is a multi-choice question, where the choices are elements on the screen. All elements are arranged in the order based on their height on the screen, from top to bottom (and from left to right). This arrangement can be used to locate them. From the screenshot, find out where and what each one is on the screen, taking into account both their text content and details. Then, determine whether one matches your target element. Please examine the choices one by one. Choose the matching one. If multiple options match your answer, choose the most likely one by re-examining the screenshot, the choices, and your further reasoning. If you would like to perform a swipe action, you can optionally select the choice where you will swipe.

A. "Home" icon
B. "Phone" icon
C. "Messages" icon
D. "Chrome" icon
E. "Search" icon
...
If none of these elements match your target element, please select Z. None of the other options match the correct element.

(Final Answer)
Finally, conclude your answer using the format below. Ensure your answer is strictly adhering to the format provided below. Please do not leave any explanation in your answers of the final standardized format part, and this final part should be clear and certain. The element choice, action, and value should be in three separate lines.

Format:

ELEMENT: The uppercase letter of your choice. (No need for TERMINATE, KEYBOARD ENTER, WAIT, ANSWER, OPEN APP, NAVIGATE HOME, NAVIGATE BACK; and optional for SWIPE.)

ACTION: Choose an action from {CLICK, INPUT TEXT, LONG PRESS, NAVIGATE BACK, TERMINATE, KEYBOARD ENTER, SWIPE, WAIT, ANSWER, OPEN APP, NAVIGATE HOME}.

VALUE: Provide additional input based on ACTION.

The VALUE means:
If ACTION == INPUT TEXT, specify the text to be typed.
If ACTION == SWIPE, specify the direction: up, down, left, right.
If ACTION == OPEN APP, provide the name of the app to be opened.
If ACTION == ANSWER, specify the text of your answer to respond directly to a question or request for information.
For CLICK, LONG PRESS, KEYBOARD ENTER, NAVIGATE HOME, NAVIGATE BACK, WAIT, and TERMINATE, write "None".
\end{lstlisting}

\section{\system Task list}
\label{sec:task_list}

The table below lists all tasks in \system. The maximum number of steps per task (``S") were determined based on human performance analysis, allowing agents approximately twice the number of steps typically required by human annotators to complete each task while preventing infinite loops.

Task completion tasks (e.g., send a message or edit a note) are abbreviated as ``TC'' and information retrieval tasks are abbreviated as ``IR''.

\small
\begin{longtable}{>{\RaggedRight\seqsplit}p{.8in}p{2in}p{.3in}p{.5in}p{.3in}p{.5in}}
\toprule
\textbf{Name} & \textbf{Template} & \textbf{Task type} & \textbf{Validation method} & \textbf{S} & \textbf{Apps} \\
\midrule
\endhead
Audio Recorder Record Audio & Record an audio clip using Audio Recorder app and save it. & TC & Filesystem & 12 & audio recorder \\
\midrule
Audio Recorder Record Audio With File Name & Record an audio clip and save it with name "\{file\_name\}" using Audio Recorder app. & TC & Filesystem & 20 & audio recorder \\
\midrule
Browser Draw & Open the file task.html in Downloads in the file manager; when prompted open it with Chrome. Then create a drawing using the three colors shown at the top and hit submit. & TC & UI\-elements & 20 & files, chrome \\
\midrule
Browser Maze & Open the file task.html in Downloads in the file manager; when prompted open it with Chrome. Then navigate the X to the bottom-right cell, by using the direction buttons. & TC & UI\-elements & 20 & files, chrome \\
\midrule
Browser Multiply & Open the file task.html in Downloads in the file manager; when prompted open it with Chrome. Then click the button 5 times, remember the numbers displayed, and enter their product in the form. & TC & UI\-elements & 22 & files, chrome \\
\midrule
Camera Take Photo & Take one photo. & TC & Filesystem & 10 & camera \\
\midrule
Camera Take Video & Take one video. & TC & Filesystem & 10 & camera \\
\midrule
Clock Stop Watch Paused Verify & Pause the stopwatch. & TC & UI\-elements & 10 & clock \\
\midrule
Clock Stop Watch Running & Run the stopwatch. & TC & UI\-elements & 10 & clock \\
\midrule
Clock Timer Entry & Create a timer with \{hours\} hours, \{minutes\} minutes, and \{seconds\} seconds. Do not start the timer. & TC & UI\-elements & 10 & clock \\
\midrule
Contacts Add Contact & Create a new contact for \{name\}. Their number is \{number\}. & TC & Database query & 12 & contacts \\
\midrule
Contacts New Contact Draft & Go to the new contact screen and enter the following details: First Name: \{first\}, Last Name: \{last\}, Phone: \{phone\}, Phone Label: \{phone\_label\}. Do NOT hit save. & TC & UI\-elements & 12 & contacts \\
\midrule
Expense Add Multiple & Add the following expenses into the pro expense: \{expense\_list\} & TC & Database query & 40 & expense \\
\midrule
Expense Add Multiple From Gallery & Add the expenses from expenses.jpg in Simple gallery to pro expense. & TC & Database query & 20 & gallery, expense \\
\midrule
Expense Add Multiple From Markor & Go through the transactions in my\_expenses.txt in Markor. Log the reimbursable transactions in the pro expense. & TC & Database query & 30 & markor, expense \\
\midrule
Expense Add Single & Add the following expenses into the pro expense: \{expense\_info\} & TC & Database query & 12 & expense \\
\midrule
Expense Delete Duplicates & Delete all but one of any expenses in pro expense that are exact duplicates, ensuring at least one instance of each unique expense remains. & TC & Database query & 12 & expense \\
\midrule
Expense Delete Duplicates2 & Delete all but one of any expenses in pro expense that are exact duplicates, ensuring at least one instance of each unique expense remains. & TC & Database query & 18 & expense \\
\midrule
Expense Delete Multiple & Delete the following expenses from pro expense: \{expense\_list\}. & TC & Database query & 20 & expense \\
\midrule
Expense Delete Multiple2 & Delete the following expenses from pro expense: \{expense\_list\}. & TC & Database query & 34 & expense \\
\midrule
Expense Delete Single & Delete the following expenses from pro expense: \{expense\_name\}. & TC & Database query & 10 & expense \\
\midrule
Files Delete File & Delete the file \{file\_name\} from the Android filesystem located in the \{subfolder\} folder within the sdk\_gphone\_x86\_64 storage area. & TC & Filesystem & 10 & files \\
\midrule
Files Move File & Move the file \{file\_name\} from \{source\_folder\} within the sdk\_gphone\_x86\_64 storage area to the \{destination\_folder\} within the same sdk\_gphone\_x86\_64 storage area in the Android filesystem. & TC & Filesystem & 20 & files \\
\midrule
Markor Add Note Header & Update the Markor note \{file\_name\} by adding the following text, along with a new blank line before the existing content: "\{header\}". & TC & Filesystem & 12 & markor \\
\midrule
Markor Change Note Content & Update the content of \{file\_name\} to "\{updated\_content\}" in Markor. & TC & Filesystem & 12 & markor \\
\midrule
Markor Create Folder & Create a new folder in Markor named \{folder\_name\}. & TC & Filesystem & 10 & markor \\
\midrule
Markor Create Note & Create a new note in Markor named \{file\_name\} with the following text: \{text\} & TC & Filesystem & 16 & markor \\
\midrule
Markor Create Note And Sms & Create a new note in Markor named \{file\_name\} with the following text: \{text\}. Share the entire content of the note with the phone number \{number\} via SMS using Simple SMS Messenger & TC & Filesystem, database query & 18 & markor, sms \\
\midrule
Markor Create Note From Clipboard & Create a note in Markor named \{file\_name\}. Perform a paste operation in the note and save the note. & TC & Filesystem & 14 & markor \\
\midrule
Markor Delete All Notes & Delete all my notes in Markor. & TC & Filesystem & 14 & markor \\
\midrule
Markor Delete Newest Note & Delete the newest note in Markor. & TC & Filesystem & 10 & markor \\
\midrule
Markor Delete Note & Delete the note in Markor named \{file\_name\}. & TC & Filesystem & 10 & markor \\
\midrule
Markor Edit Note & Edit \{file\_name\} in Markor. \{edit\_subcommand\} & TC & Filesystem & 12 & markor \\
\midrule
Markor Merge Notes & Merge the contents of Markor notes \{file1\_name\}, \{file2\_name\} and \{file3\_name\} (in the same order) into a new Markor note named \{new\_file\_name\} and save it. Add a new line between the content of each note. & TC & Filesystem & 78 & markor \\
\midrule
Markor Move Note & In Markor, move the note \{file\_name\} from \{source\_folder\} to \{destination\_folder\}. & TC & Filesystem & 14 & markor \\
\midrule
Markor Transcribe Receipt & Create a file in Markor, called receipt.md with the transactions from the receipt.png. Use Simple Gallery to view the receipt. Please enter transactions in csv format including the header "Date, Item, Amount". & TC & Filesystem & 18 & gallery, markor \\
\midrule
Markor Transcribe Video & Transcribe the contents of video \{video\_name\} by watching it in VLC player (located in Download) and writing the sequence of strings shown on each frame to the text file \{file\_name\} in Markor as a comma separated list. For example, if the first frame shows the text "edna" and the second frame shows the text "pineapple", then the text file should contains only the following text: "edna, pineapple". & TC & Filesystem & 20 & markor, vlc \\
\midrule
Notes Is Todo & Is the note titled '\{title\}' in the Joplin app marked as a todo item? Respond with either 'True' if it is a todo or 'False' if not. & IR & String match & 10 & joplin \\
\midrule
Notes Meeting Attendee Count & How many attendees were present in the meeting titled '\{title\}' in the Joplin app? Express your answer as just a single number. & IR & String match & 10 & joplin \\
\midrule
Notes Recipe Ingredient Count & What quantity of \{ingredient\} do I need for the recipe `\{title\}' in the Joplin app? Express your answer in the format $\langle$amount$\rangle$ $\langle$unit$\rangle$ without using abbreviations. & IR & String match & 10 & joplin \\
\midrule
Notes Todo Item Count & How many to-dos do I have in the '\{folder\}' folder in the Joplin app? Express your answer as just a single number. & IR & String match & 10 & joplin \\
\midrule
Open App Task Eval & Open the \{app\_name\} app. Clear any pop-ups that may appear by granting all permissions that are required. & TC & System API & 10 & camera, clock, contacts, settings, dialer \\
\midrule
Osm And Favorite & Add a favorite location marker for \{location\} in the OsmAnd maps app. & TC & Filesystem & 13 & osmand \\
\midrule
Osm And Marker & Add a location marker for \{location\} in the OsmAnd maps app. & TC & Filesystem & 20 & osmand \\
\midrule
Osm And Track & Save a track with waypoints Ruggell, Liechtenstein, Bendern, Liechtenstein in the OsmAnd maps app in the same order as listed. & TC & Filesystem & 120 & osmand \\
\midrule
Recipe Add Multiple Recipes & Add the following recipes into the Broccoli app: \{recipe\_list\} & TC & Database query & 68 & recipe \\
\midrule
Recipe Add Multiple Recipes From Image & Add the recipes from recipes.jpg in Simple gallery to the Broccoli recipe app. & TC & Database query & 26 & markor, recipe \\
\midrule
Recipe Add Multiple Recipes From Markor & Add the recipes from recipes.txt in Markor to the Broccoli recipe app. & TC & Database query & 48 & gallery, recipe \\
\midrule
Recipe Add Multiple Recipes From Markor2 & Add the recipes from recipes.txt in Markor that take 10 mins to prepare into the Broccoli recipe app. & TC & Database query & 52 & recipe \\
\midrule
Recipe Add Single Recipe & Add the following recipes into the Broccoli app: \{recipe\_list\} & TC & Database query & 24 & recipe \\
\midrule
Recipe Delete Duplicate Recipes & Delete all but one of any recipes in the Broccoli app that are exact duplicates, ensuring at least one instance of each unique recipe remains & TC & Database query & 10 & recipe \\
\midrule
Recipe Delete Duplicate Recipes2 & Delete all but one of any recipes in the Broccoli app that are exact duplicates, ensuring at least one instance of each unique recipe remains & TC & Database query & 24 & recipe \\
\midrule
Recipe Delete Duplicate Recipes3 & Delete all but one of any recipes in the Broccoli app that are exact duplicates, ensuring at least one instance of each unique recipe remains & TC & Database query & 34 & recipe \\
\midrule
Recipe Delete Multiple Recipes & Delete the following recipes from Broccoli app: \{recipe\_list\} & TC & Database query & 24 & recipe \\
\midrule
Recipe Delete Multiple Recipes With Constraint & Delete the recipes from Broccoli app that use \{ingredient\} in the directions. & TC & Database query & 40 & recipe \\
\midrule
Recipe Delete Multiple Recipes With Noise & Delete the following recipes from Broccoli app: \{recipe\_list\} & TC & Database query & 34 & recipe \\
\midrule
Recipe Delete Single Recipe & Delete the following recipes from Broccoli app: \{recipe\_list\} & TC & Database query & 10 & recipe \\
\midrule
Recipe Delete Single With Recipe With Noise & Delete the following recipes from Broccoli app: \{recipe\_list\} & TC & Database query & 20 & recipe \\
\midrule
Retro Create Playlist & Create a playlist in Retro Music titled "\{title\}" with the following songs, in order: \{song\_list\} & TC & Database query & 24 & music \\
\midrule
Retro Playing Queue & Add the following songs, in order, \{song\_list\} to my playing queue in Retro music. & TC & Database query & 32 & music \\
\midrule
Retro Playlist Duration & Create a playlist in Retro Music titled "\{title\}" with a duration between 45 and 50 minutes using the provided songs. & TC & Database query & 30 & music \\
\midrule
Retro Save Playlist & Create a playlist in Retro Music titled "\{title\}" with the following songs, in order: \{song\_list\}. Then export the playlist to the Downloads directory on the device. & TC & Database query & 50 & music \\
\midrule
Save Copy Of Receipt Task Eval & Copy \{file\_name\} in DCIM and save a copy with the same name in Download & TC & Filesystem & 16 & gallery \\
\midrule
Simple Calendar Add One Event & In Simple calendar, create a calendar event on \{year\}-\{month\}-\{day\} at \{hour\}h with the title '\{event\_title\}' and the description '\{event\_description\}'. The event should last for \{duration\_mins\} mins. & TC & Database query & 34 & calendar \\
\midrule
Simple Calendar Add One Event In Two Weeks & In Simple calendar, create a calendar event in two weeks from today at \{hour\}h with the title '\{event\_title\}' and the description '\{event\_description\}'. The event should last for \{duration\_mins\} mins. & TC & Database query & 20 & calendar \\
\midrule
Simple Calendar Add One Event Relative Day & In Simple calendar, create a calendar event for this \{day\_of\_week\} at \{hour\}h with the title '\{event\_title\}' and the description '\{event\_description\}'. The event should last for \{duration\_mins\} mins. & TC & Database query & 34 & calendar \\
\midrule
Simple Calendar Add One Event Tomorrow & In Simple calendar, create a calendar event for tomorrow at \{hour\}h with the title '\{event\_title\}' and the description '\{event\_description\}'. The event should last for \{duration\_mins\} mins. & TC & Database query & 26 & calendar \\
\midrule
Simple Calendar Add Repeating Event & In Simple calendar, create a recurring calendar event titled '\{event\_title\}' starting on \{year\}-\{month\}-\{day\} at \{hour\}h. The event recurs \{repeat\_rule\}, forever, and lasts for \{duration\_mins\} minutes each occurrence. The event description should be '\{event\_description\}'. & TC & Database query & 28 & calendar \\
\midrule
Simple Calendar Any Events On Date & Do I have any events \{date\} in Simple calendar? Answer with the titles only. If there are multiples titles, format your answer in a comma separated list. & IR & Database query & 10 & calendar \\
\midrule
Simple Calendar Delete Events & In Simple calendar, delete all the calendar events on \{year\}-\{month\}-\{day\} & TC & Database query & 14 & calendar \\
\midrule
Simple Calendar Delete Events On Relative Day & In Simple calendar, delete all events scheduled for this \{day\_of\_week\}. & TC & Database query & 12 & calendar \\
\midrule
Simple Calendar Delete One Event & In Simple calendar, delete the calendar event on \{year\}-\{month\}-\{day\} at \{hour\}h with the title '\{event\_title\}' & TC & Database query & 12 & calendar \\
\midrule
Simple Calendar Event On Date At Time & What is on my schedule for \{date\} at \{time\} in Simple calendar? Answer with the titles only. If there are multiples titles, format your answer in a comma separated list. & IR & Database query & 10 & calendar \\
\midrule
Simple Calendar Events In Next Week & What events do I have in the next week in Simple calendar? Answer with the titles only. If there are multiples titles, format your answer in a comma separated list. & IR & Database query & 10 & calendar \\
\midrule
Simple Calendar Events In Time Range & Do I have any events between \{start\_time\} and 8pm \{date\} in Simple calendar? Answer with the titles only. If there are multiples titles, format your answer in a comma separated list. & IR & Database query & 10 & calendar \\
\midrule
Simple Calendar Events On Date & What events do I have \{date\} in Simple calendar? Answer with the titles only. If there are multiple titles, format your answer as a comma separated list. & IR & Database query & 10 & calendar \\
\midrule
Simple Calendar First Event After Start Time & What is my first event after \{time\} \{date\} in Simple calendar? Answer with the titles only. If there are multiples titles, format your answer in a comma separated list. & IR & Database query & 10 & calendar \\
\midrule
Simple Calendar Location Of Event & What is the location of my \{title\} event in Simple calendar? Answer with the location only. & IR & Database query & 10 & calendar \\
\midrule
Simple Calendar Next Event & What is my next upcoming event in Simple calendar? Answer with the title only. If there are multiples titles, format your answer in a comma separated list. & IR & Database query & 10 & calendar \\
\midrule
Simple Calendar Next Meeting With Person & When is my next meeting with \{person\} in Simple calendar? Express your answer in the format $\langle$month name$\rangle$ $\langle$day$\rangle$ $\langle$year$\rangle$ $\langle$hour in 24-hour format$\rangle$:$\langle$minutes$\rangle$. & IR & Database query & 10 & calendar \\
\midrule
Simple Draw Pro Create Drawing & Create a new drawing in Simple Draw Pro. Name it \{file\_name\}. Save it in the Pictures folder within the sdk\_gphone\_x86\_64 storage area. & TC & Filesystem & 18 & simpledrawpro \\
\midrule
Simple Sms Reply & Reply to \{number\} with message: \{message\} in Simple SMS Messenger & TC & Database query & 12 & sms \\
\midrule
Simple Sms Reply Most Recent & Reply to the most recent text message using Simple SMS Messenger with message: \{message\} & TC & Database query & 12 & sms \\
\midrule
Simple Sms Resend & Resend the message I just sent to \{name\} in Simple SMS Messenger & TC & Database query & 12 & sms \\
\midrule
Simple Sms Send & Send a text message using Simple SMS Messenger to \{number\} with message: \{message\} & TC & Database query & 12 & sms \\
\midrule
Simple Sms Send Clipboard Content & Send a message to \{number\} with the clipboard content in Simple SMS Messenger & TC & Database query & 12 & sms \\
\midrule
Simple Sms Send Received Address & Text the address of the event to \{name1\} that \{name2\} just sent me in Simple SMS Messenger & TC & Database query & 18 & sms \\
\midrule
Sports Tracker Activities Count For Week & How many \{category\} activities did I do this week in the OpenTracks app? Express your answer as a single integer. & IR & String match & 10 & sportstracker \\
\midrule
Sports Tracker Activities On Date & What activities did I do \{date\} in the OpenTracks app? Answer with the category only. If there are multiples categories, format your answer in a comma separated list. & IR & String match & 20 & sportstracker \\
\midrule
Sports Tracker Activity Duration & How long was my \{category\} activity \{date\} in the OpenTracks app? Express your answer in minutes as a single integer. & IR & String match & 12 & sportstracker \\
\midrule
Sports Tracker Longest Distance Activity & What was the longest distance covered in a \{category\} activity in the OpenTracks app this week? Express your answer in meters as a single integer. & IR & String match & 10 & sportstracker \\
\midrule
Sports Tracker Total Distance For Category Over Interval & What was the total distance covered for \{category\} activities in the OpenTracks app from \{start\_date\} to \{end\_date\}? Express your answer in meters as a single integer. & IR & String match & 22 & sportstracker \\
\midrule
Sports Tracker Total Duration For Category This Week & What was the total duration of \{category\} activities in the OpenTracks app this week? Express your answer in minutes as a single integer. & IR & String match & 16 & sportstracker \\
\midrule
System Bluetooth Turn Off & Turn bluetooth off. & TC & System API & 10 & settings \\
\midrule
System Bluetooth Turn Off Verify & Turn bluetooth off. & TC & System API & 10 & settings \\
\midrule
System Bluetooth Turn On & Turn bluetooth on. & TC & System API & 10 & settings \\
\midrule
System Bluetooth Turn On Verify & Turn bluetooth on. & TC & System API & 10 & settings \\
\midrule
System Brightness Max & Turn brightness to the max value. & TC & System API & 10 & settings \\
\midrule
System Brightness Max Verify & Turn brightness to the max value. & TC & System API & 10 & settings \\
\midrule
System Brightness Min & Turn brightness to the max value. & TC & System API & 10 & settings \\
\midrule
System Brightness Min Verify & Turn brightness to the max value. & TC & System API & 10 & settings \\
\midrule
System Copy To Clipboard & Copy the following text to the clipboard: \{clipboard\_content\} & TC & System API & 10 & n/a \\
\midrule
System Wifi Turn Off & Turn wifi off. & TC & System API & 10 & settings \\
\midrule
System Wifi Turn Off Verify & Turn wifi off. & TC & System API & 10 & settings \\
\midrule
System Wifi Turn On & Turn wifi on. & TC & System API & 10 & settings \\
\midrule
System Wifi Turn On Verify & Turn wifi on. & TC & System API & 10 & settings \\
\midrule
Tasks Completed Tasks For Date & Which tasks have I completed for \{date\} in Tasks app? Answer with the titles only. If there are multiples titles, format your answer in a comma separated list. & IR & String match & 10 & tasks \\
\midrule
Tasks Due Next Week & How many tasks do I have due next week in Tasks app? Express your answer as a single integer. & IR & String match & 12 & tasks \\
\midrule
Tasks Due On Date & What tasks do I have due \{date\} in Tasks app? Answer with the titles only. If there are multiples titles, format your answer in a comma separated list. & IR & String match & 10 & tasks \\
\midrule
Tasks High Priority Tasks & What are my high priority tasks in Tasks app? Answer with the titles only. If there are multiples titles, format your answer in a comma separated list. & IR & String match & 10 & tasks \\
\midrule
Tasks High Priority Tasks Due On Date & Which tasks with high priority are due \{date\} in the Tasks app? Answer with the title only. If there are multiples titles, format your answer in a comma separated list. & IR & String match & 10 & tasks \\
\midrule
Tasks Incomplete Tasks On Date & What incomplete tasks do I have still have to do by \{date\} in Tasks app? Answer with the titles only. If there are multiples titles, format your answer in a comma separated list. & IR & String match & 10 & tasks \\
\midrule
Turn Off Wifi And Turn On Bluetooth & Turn off WiFi, then enable bluetooth & TC & String match & 20 & settings \\
\midrule
Turn On Wifi And Open App & Turn on Wifi, then open the \{app\_name\} app & TC & String match & 20 & settings \\
\midrule
Vlc Create Playlist & Create a playlist titled \{title\}" with the following files in VLC (located in Internal Memory/VLCVideos), in order: \{video\_names\} & TC & String match & 28 & vlc \\
\midrule
Vlc Create Two Playlists & Create a playlist titled "\{title1\}" with the following files in VLC (located in Internal Memory/VLCVideos), in order: \{video\_names1\}. And then, create a playlist titled "\{title2\}" with the following files in VLC, in order: \{video\_names2\}. & TC & String match & 48 & vlc \\
\bottomrule
\label{tab:all-task-list}
\end{longtable}

\end{appendices}

\end{document}